\documentclass{article}

\usepackage[final]{neurips_2023}
\usepackage{graphicx}
\usepackage{subfigure}
\usepackage{wrapfig}

\usepackage{amsmath}
\usepackage{amssymb}
\usepackage{mathtools}
\usepackage{amsthm}

\usepackage[utf8]{inputenc} 
\usepackage[T1]{fontenc}    
\usepackage{hyperref}       
\usepackage{url}            
\usepackage{booktabs}       
\usepackage{amsfonts}       
\usepackage{nicefrac}       
\usepackage{microtype}      
\usepackage{xcolor}         
\usepackage{footmisc}

\usepackage{tcolorbox}

\hypersetup{colorlinks,linkcolor={blue},citecolor={magenta},urlcolor={blue}}  
\usepackage[capitalize,noabbrev]{cleveref}

\title{Small batch deep reinforcement learning}

\author{%
  Johan Obando-Ceron \thanks{Work done during an internship at Google DeepMind} \\
  Mila, Université de Montréal \\
  \texttt{jobando0730@gmail.com} \\
  \And
  Marc G. Bellemare \\
  Mila, Université de Montréal\\
  \texttt{bellemam@mila.quebec}
  \And
  Pablo Samuel Castro \\
  Google DeepMind \\
  Mila, Université de Montréal \\
  \texttt{psc@google.com} \\
}

\begin{document}

\maketitle

\begin{abstract}
In value-based deep reinforcement learning with replay memories, the batch size parameter specifies how many transitions to sample for each gradient update. Although critical to the learning process, this value is typically not adjusted when proposing new algorithms. In this work we present a broad empirical study that suggests {\em reducing} the batch size can result in a number of significant performance gains; this is surprising, as the general tendency when training neural networks is towards larger batch sizes for improved performance. We complement our experimental findings with a set of empirical analyses towards better understanding this phenomenon.

\end{abstract}
\section{Introduction}
\label{sec:intro}

One of the central concerns for deep reinforcement learning (RL) is how to efficiently make the most use of the collected data for policy improvement. This is particularly important in online settings, where RL agents learn while interacting with an environment, as interactions can be expensive. Since the introduction of DQN \citep{mnih2015humanlevel}, one of the core components of most modern deep RL algorithms is the use of a finite {\em replay memory} where experienced transitions are stored. During learning, the agent samples mini-batches from this memory to update its network parameters.

Since the policy used to collect transitions is changing throughout learning, the replay memory contains data coming from a mixture of policies (that differ from the agent's current policy), and results in what is known as {\em off-policy} learning. In contrast with training data for supervised learning problems, online RL data is highly {\em non-stationary}. Still, at any point during training the replay memory exhibits a distribution over transitions, which the agent samples from at each learning step. The number of sampled transitions at each learning step is known as the {\em batch size}, and is meant to produce an unbiased estimator of the underlying data distribution. Thus, in theory, larger batch sizes should be more accurate representations of the true distribution.

Some in the supervised learning community suggest that learning with large batch sizes leads to better optimization \citep{shallue19measuring}, since smaller batches yield noisier gradient estimations. Contrastingly, others have observed that larger batch sizes tend to converge to ``sharper'' optimization landscapes, which can result in worsened generalization \citep{keskar2017on}; smaller batches, on the other hand, seem to result in ``flatter'' landscapes, resulting in better generalization.

Learning dynamics in deep RL are drastically different than those observed in supervised learning, in large part due to the data non-stationarity mentioned above. Given that the choice of batch size will have a direct influence on the agent's sample efficiency and ultimate performance, developing a better understanding of its impact is critical. Surprisingly, to the best of our knowledge there have been no studies exploring the impact of the choice of batch size in deep RL. Most recent works have focused on related questions, such as the number of gradient updates per environment step \citep{nikishin22primacy,d'oro2023sampleefficient,sokar23redo}, but have kept the batch size fixed.

In this work we conduct a broad empirical study of batch size in online value-based deep reinforcement learning. We uncover the surprising finding that {\em reducing} the batch size seems to provide substantial performance benefits and computational savings. We showcase this finding in a variety of agents and training regimes (\autoref{sec:varyingbatchsize_effects}), and conduct in-depth analyses of the possible causes (\autoref{sec:potential_causes}). The impact of our findings and analyses go beyond the choice of the batch size hyper-parameter, and help us develop a better understanding of the learning dynamics in online deep RL.
\section{Background}
\label{sec:brackground}

\begin{figure}[!t]
    \centering
  \includegraphics[width=\linewidth]{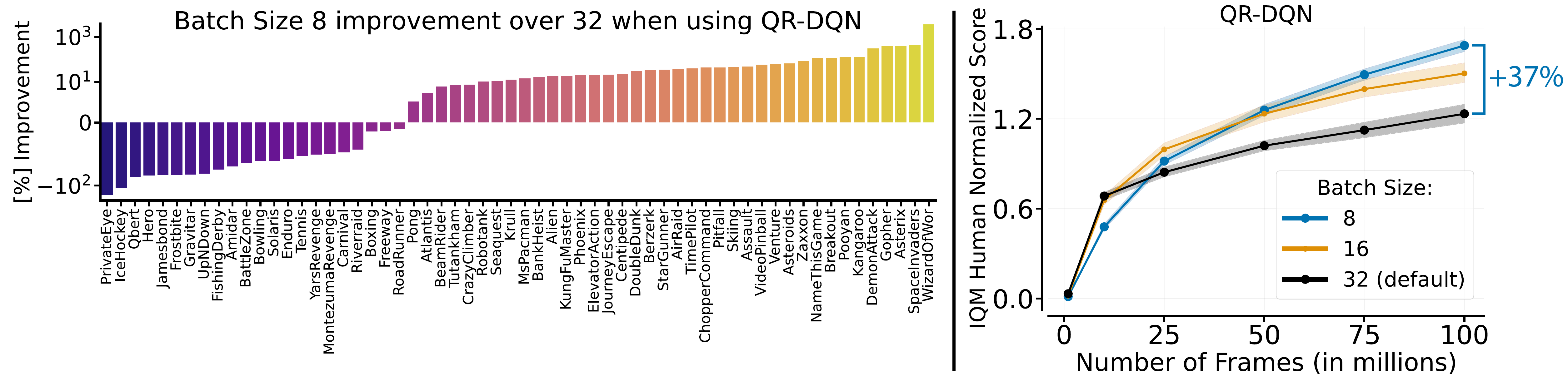}
    \caption{Evaluating QR-DQN \citep{dabney18distributional} with varying batch sizes over all 60 Atari 2600 games. {\bf (Left)} Average improvement obtained when using a batch size of 8 over 32 (default); {\bf (Right)} Aggregate Interquantile Mean \citep{agarwal2021deep} of human normalized scores. All games run for 3 seeds, with shaded areas displaying 95\% stratified bootstrap confidence intervals.}%
    \label{fig:all_games}%
    \vspace{-1em}
\end{figure}

A reinforcement learning problem is typically formulated as a Markov decision process (MDP), which consists of a 5-tuple $\langle \mathcal{S}, \mathcal{A}, \mathcal{P}, \mathcal{R}, \gamma, \rangle$, where $\mathcal{S}$ denotes the state space, $\mathcal{A}$ denotes the actions, $\mathcal{P}:\mathcal{S}\times\mathcal{A}\rightarrow Dist(\mathcal{S})$ encodes the transition dynamics, $\mathcal{R}:\mathcal{S}\times \mathcal{A}\rightarrow\mathbb{R}$ is the reward function, and $\gamma\in[0, 1)$ is a discount factor. The aim is to learn a {\em policy} $\pi_\theta: \mathcal{S} \mapsto \mathcal{A}$ parameterized by $\theta$ such that the sum of discounted returns $\mathbb{E}_{\pi_\theta}\left[\sum_{t=1}^{\infty} \gamma^t r_t\right]$ is maximized; here, the state-action trajectory $\left(\mathbf{s}_0, \mathbf{a}_0, \mathbf{s}_1, \mathbf{a}_1, \ldots\right)$ is obtained by sampling an action $\mathbf{a}_t \sim \pi_\theta\left(\cdot \mid \mathbf{s}_t\right)$ and reaching state $\mathbf{s}_{t+1} \sim \mathcal{P}\left(\cdot \mid \mathbf{s}_t, \mathbf{a}_t\right)$ at each decision step $t$, and $r_t \sim \mathcal{R}\left(\cdot \mid \mathbf{s}_t, \mathbf{a}_t\right)$. 

In value-based methods, the policy is obtained as the argmax of a learned $Q$-function: $\pi_\theta(s)\equiv\arg\max_{a\in\mathcal{A}}Q_{\theta}(s,a)$. This function aims to approximate the optimal state-action values $Q^*$, defined via the well-known Bellman recurrence: $Q^*(\mathbf{s}_t, \mathbf{a}_t)=\max _{\mathbf{a}^{\prime}} \mathbb{E}[\mathcal{R}(\mathbf{s}_t, \mathbf{a}_t)+$ $\left.\gamma Q^*\left(\mathbf{s}_{t+1}, \mathbf{a}_{t+1}\right)\right]$, and is typically learned using $Q$-learning \citep{watkins1992q,sutton2018reinforcement}.

To deal with large state spaces, such as all possible images in an Atari 2600 game, \citet{mnih2015humanlevel} introduced DQN, which combined Q-learning with deep neural networks to represent $Q_{\theta}$. A large {\em replay buffer} $D$ is maintained to store experienced transitions, from which mini-batches are sampled to perform learning updates \citep{lin92self}. Specifically, {\em temporal difference learning} is used to update the network parameters with the following loss function: $L(\theta) = \mathbb{E}_{(s_t, a_t, r_t, s_{t+1})\sim D}[(\left( r_{t} + \gamma\max_{a'\in\mathcal{A}}Q_{\bar{\theta}}(s_{t+1}, a_{t+1})\right) - Q_{\theta}(s_t, a_t))^2]$. Here $Q_{\bar{\theta}}$ is a {\em target network} that is a delayed copy of $Q_{\theta}$, with the parameters synced with $Q_{\theta}$ less frequently than $Q_{\theta}$ is updated.

Since the introduction of DQN, there have been a number of algorithmic advances in deep RL agents, in particular those which make use of distributional RL \citep{bellemare17distributional}, introduced with the C51 algorithm. The Rainbow agent combined C51 with other advances such as multi-step learning and prioritized replay sampling \citep{hessel18rainbow}. Different ways of parameterizing return distributions were proposed in the form of the IQN \citep{dabney18iqn} and QR-DQN \citep{dabney18distributional} algorithms. For reasons which will be clarified below, most of our evaluations and analyses were conducted with the QR-DQN agent.

\section{The small batch effect on agent performance}
\label{sec:varyingbatchsize_effects}
In this section we showcase the performance gains that arise when training with smaller batch sizes. We do so first with four standard value-based agents (\S\ref{sec:onlineTraining}), with varying architectures (\S\ref{sec:architectures}), agents optimized for sample efficiency (\S\ref{sec:lowDataRegime}), and with extended training (\S\ref{sec:trainingStability}). Additionally, we explore the impact of reduced batch sizes on exploration (\S\ref{sec:exploration}) and computational cost (\S\ref{sec:computationalConsequences}).

\noindent {\bf Experimental setup:}
We use the Jax implementations of RL agents, with their default hyper-parameter values, provided by the Dopamine library \citep{castro18dopamine}\footnote{Dopamine code available at https://github.com/google/dopamine.} and applied to the Arcade Learning Environment (ALE) \citep{Bellemare_2013}.\footnote{Dopamine uses sticky actions by default \citep{revisiting_arcade}.} It is worth noting that the default batch size is $32$, which we indicate with a {\bf black} color in all the plots below, for clarity. We evaluate our agents on 20 games chosen by \citet{fedus2020revisiting} for their analysis of replay ratios, picked to offer a diversity of difficulty and dynamics. To reduce the computational burden, we ran most of our experiments for 100 million frames (as opposed to the standard 200 million).
For evaluation, we follow the guidelines of \citet{agarwal2021deep}. Specifically, we run 3 independent seeds for each experiment and report the human-normalized {\em interquantile mean (IQM)}, aggregated over the 20 games, configurations, and seeds, with the 95\% stratified bootstrap confidence intervals. Note that this means that for most of the aggregate results presented here, we are reporting mean and confidence intervals over 60 independent seeds. All experiments were run on NVIDIA Tesla P100 GPUs.

\subsection{Standard agents}
\label{sec:onlineTraining}

\begin{wrapfigure}{r}{0.47\textwidth}
   \vspace{-1em}
    \centering
    \includegraphics[width=1\linewidth]{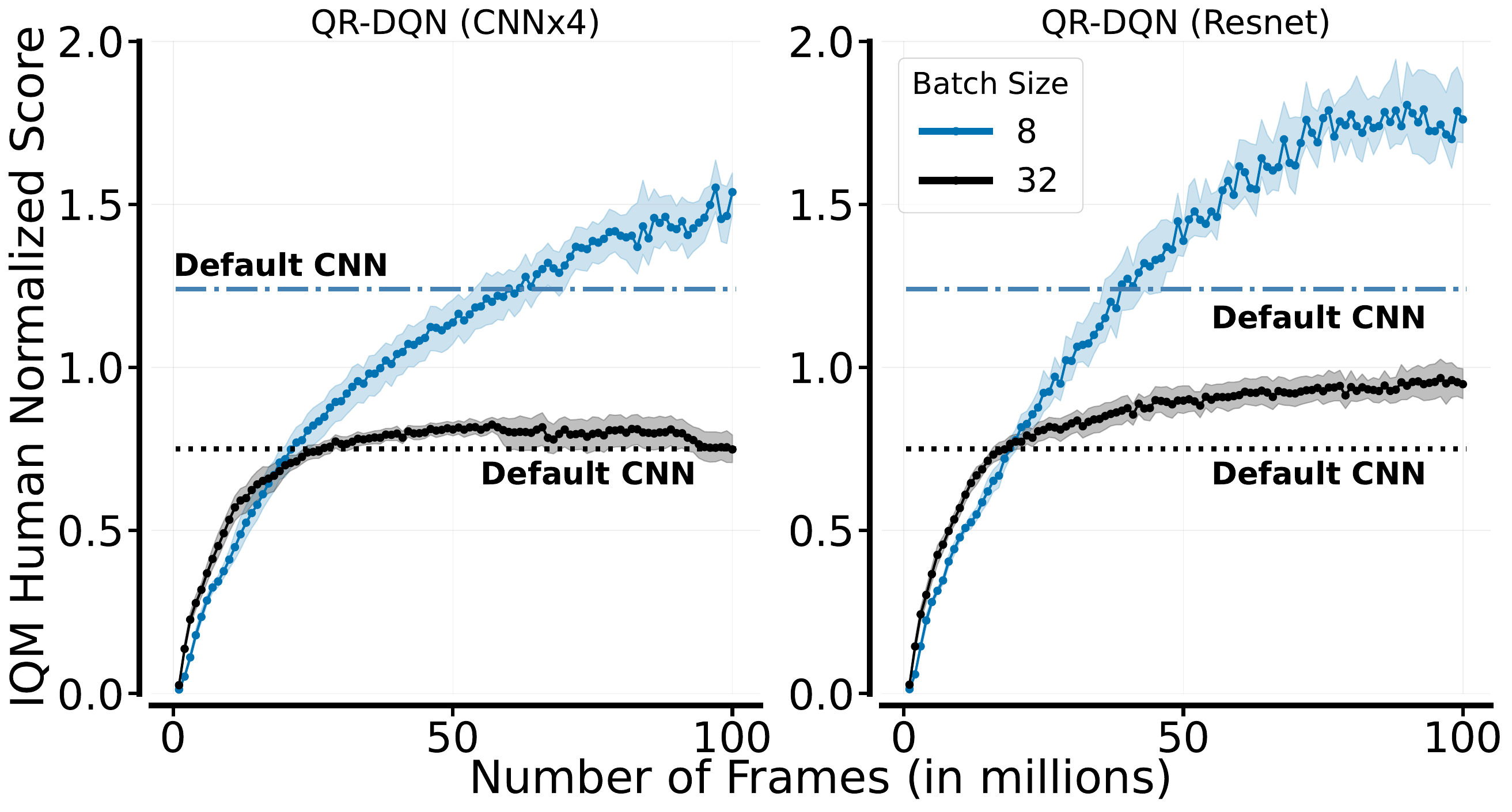}%
    \caption{IQM for human normalized scores with varying neural network architectures over 20 games, with 3 seeds per experiment. Shaded areas represent 95\% stratified bootstrap confidence intervals.}%
     \vspace{-1em}
    \label{fig:architectureResults}%
\end{wrapfigure}

\begin{figure}[!t]
    \centering
   \includegraphics[width=0.248\textwidth]{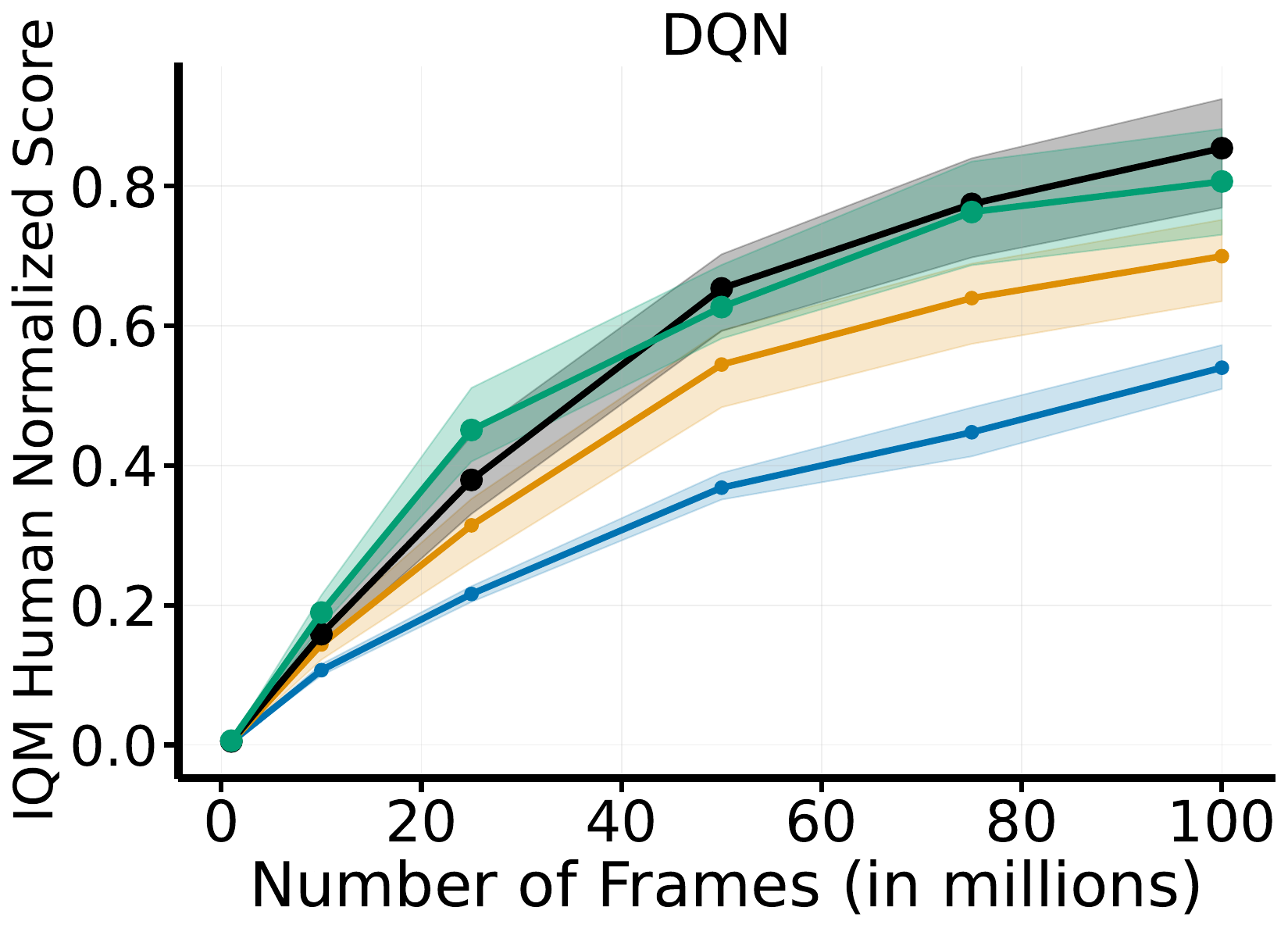}%
   \includegraphics[width=0.248\textwidth]{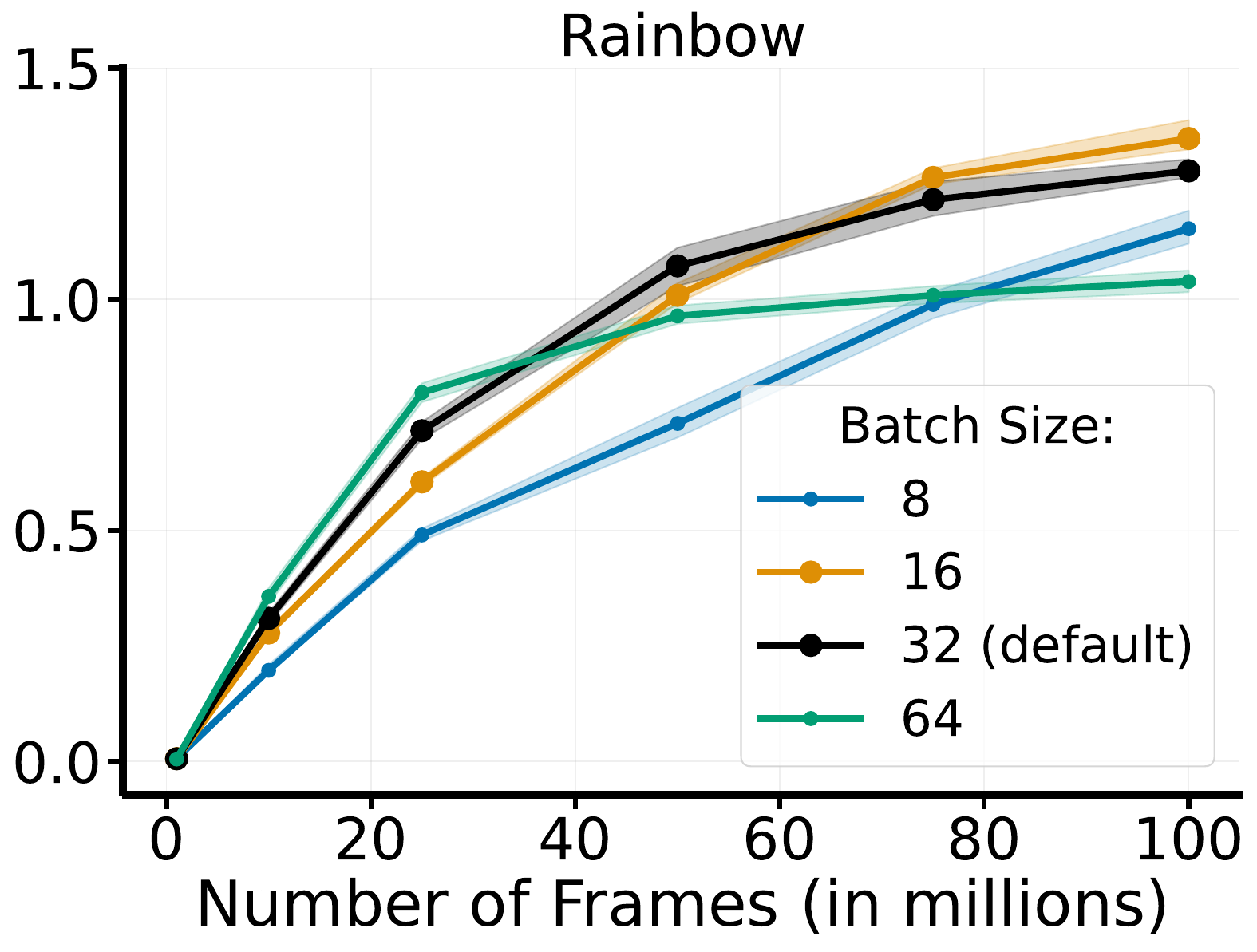}
   \includegraphics[width=0.248\textwidth]{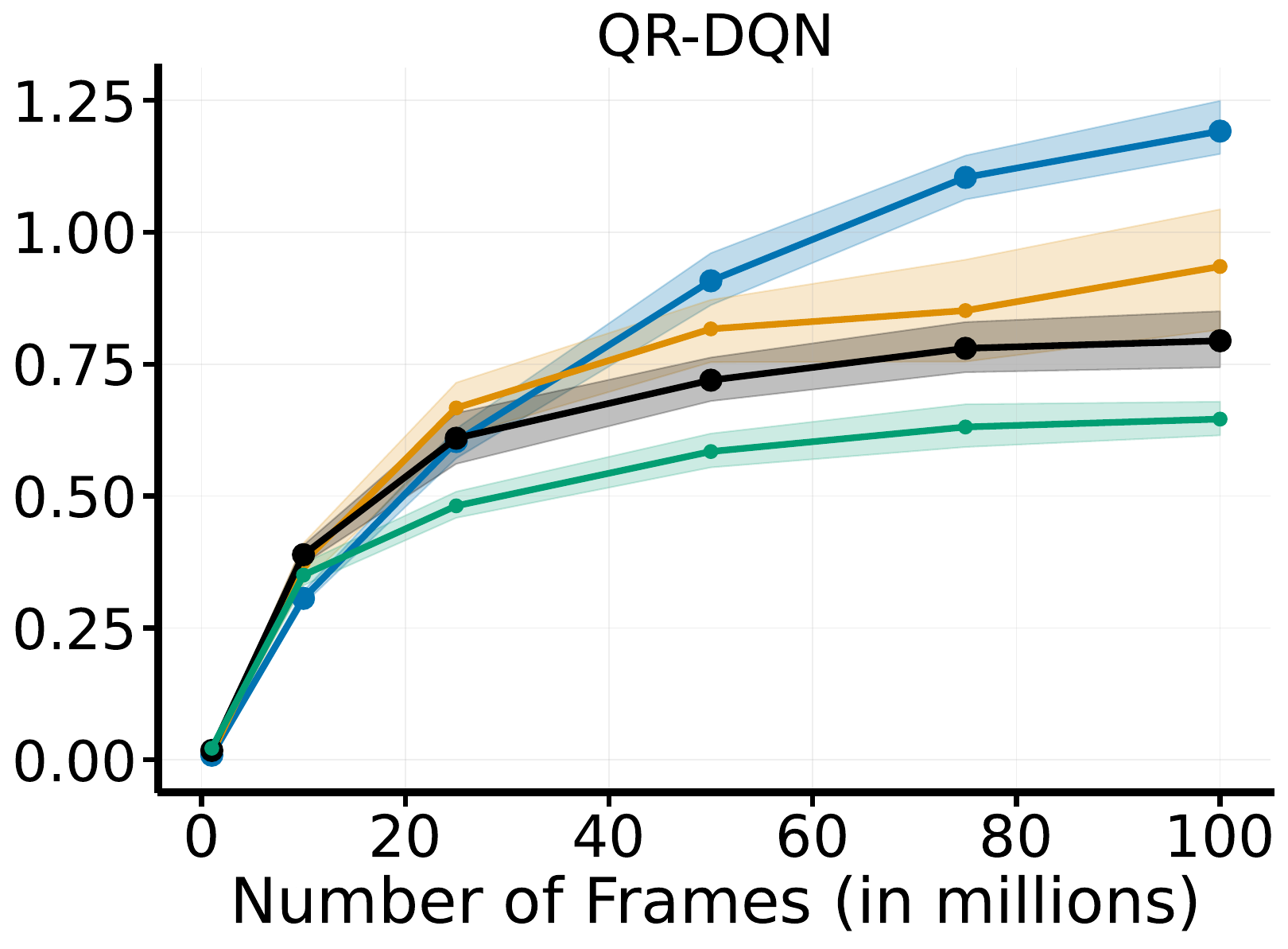}%
   \includegraphics[width=0.248\textwidth]{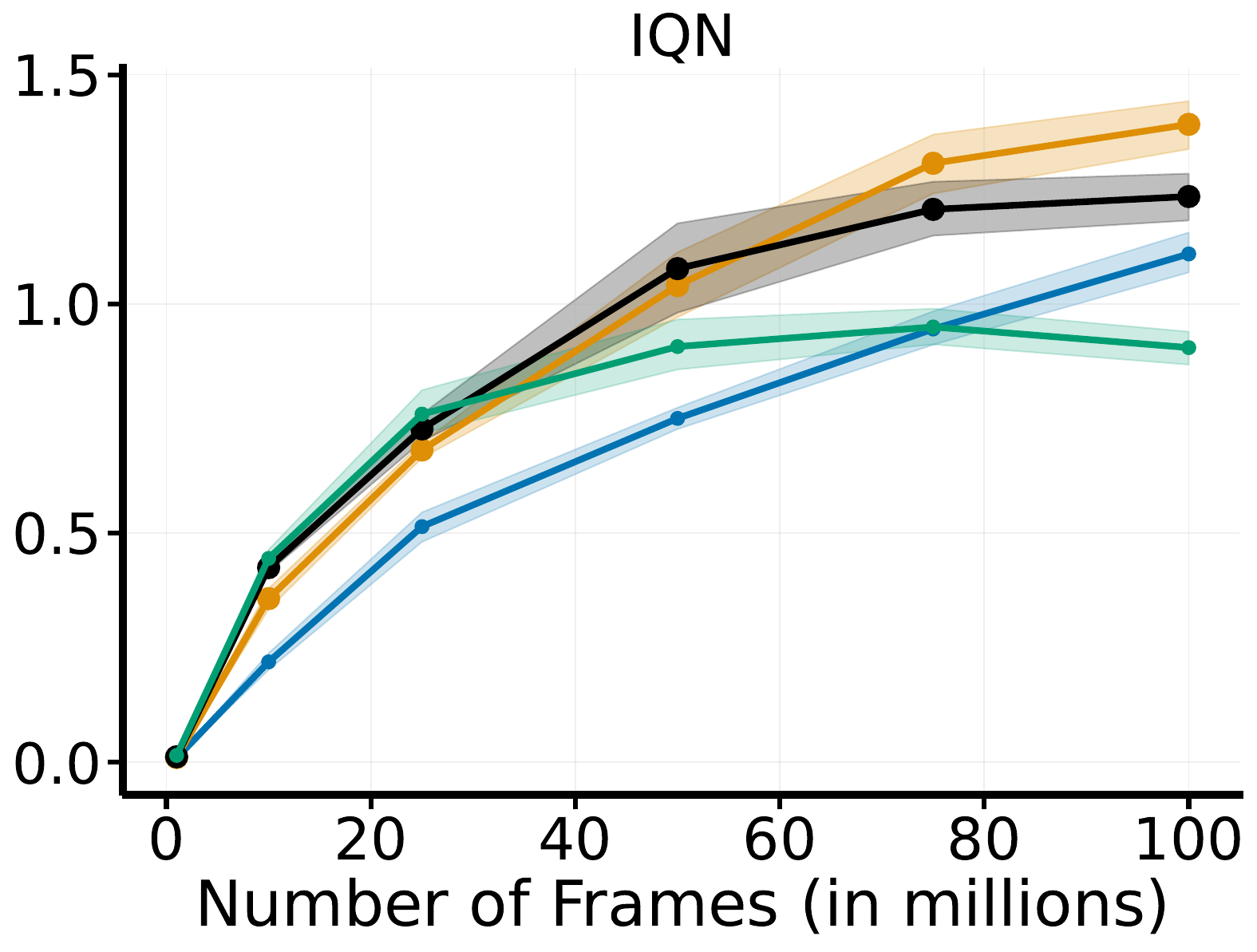}%
    \caption{IQM for human normalized scores for DQN, Rainbow, QR-DQN,  and IQN over 20 games. All games run with 3 independent seeds, shaded areas representing 95\% confidence intervals.}
    \label{fig:iqmCurveDefaultDQNRainbowQRDQNIQN}
    \vspace{-1em}
\end{figure}

We begin by investigating the impact reducing the batch size can have on four popular value-based agents, which were initially benchmarked on the ALE suite: DQN \citep{mnih2015humanlevel}, Rainbow \citep{hessel18rainbow} (Note that Dopamine uses a ``compact'' version of the original Rainbow agent, including only multi-step updates, prioritized replay, and C51), QR-DQN \citep{dabney18distributional}, and IQN \citep{dabney18iqn}. In \autoref{fig:iqmCurveDefaultDQNRainbowQRDQNIQN} we can observe that, in general, reduced batch size results in improved performance. The notable exception is DQN, for which we provide an analysis and explanation for why this is the case below. To verify that our results are not a consequence of the set of 20 games used in our analyses, we ran QR-DQN (where the effect is most observed) over the full 60 games in the suite and report the results in \autoref{fig:all_games}. Remarkably, a batch size of 8 results in significant gains on 38 out of the full 60 games, for an average performance improvement of 98.25\%.

\subsection{Varying architectures} 
\label{sec:architectures}

Although the CNN architecture originally introduced by DQN \citep{mnih2015humanlevel} has been the backbone for most deep RL networks, there have been some recent works exploring the effects of varying architectures \citep{espeholt18impala,agarwal2022beyond,sokar23redo}. We investigate the small batch effect by varying the QR-DQN architecture in two ways: {\bf (1)} expanding the convolutional widths by 4 times (resulting in a substantial increase in the number of parameters), and {\bf (2)} using the Resnet architecture proposed by \citet{espeholt18impala} (which results in a similar number of parameters to the original CNN architecture, but is a deeper network). In \autoref{fig:architectureResults} we can observe that not only do reduced batch sizes yield improved performance, but they are better able to leverage the increased number of parameters (CNNx4) and the increased depth (Resnet).

\subsection{Atari 100k agents}
\label{sec:lowDataRegime}
There has been an increased interest in evaluating Atari agents on very few environment interactions, for which \citet{Kaiser2020Model} proposed the 100k benchmark\footnote{Here, 100k refers to agent steps, or 400k environment frames, due to skipping frames in the standard training setup.}. We evaluate the effect of reduced batch size on three of the most widely used agents for this regime: 
Data-efficient Rainbow (DER), a version of the Rainbow algorithm with hyper-parameters tuned for faster early learning \citep{DER}; DrQ($ \epsilon $), which is a variant of DQN that uses data augmentation \citep{agarwal2021deep}; and
SPR, which incorporates self-supervised learning to improve sample efficiency \citep{schwarzer2020data}.
For this evaluation we evaluate on the standard 26 games for this benchmark \citep{Kaiser2020Model}, aggregated over 6 independent trials.

\begin{figure*}[!t]
    \centering
    \vfill
    \includegraphics[width=4.6cm]{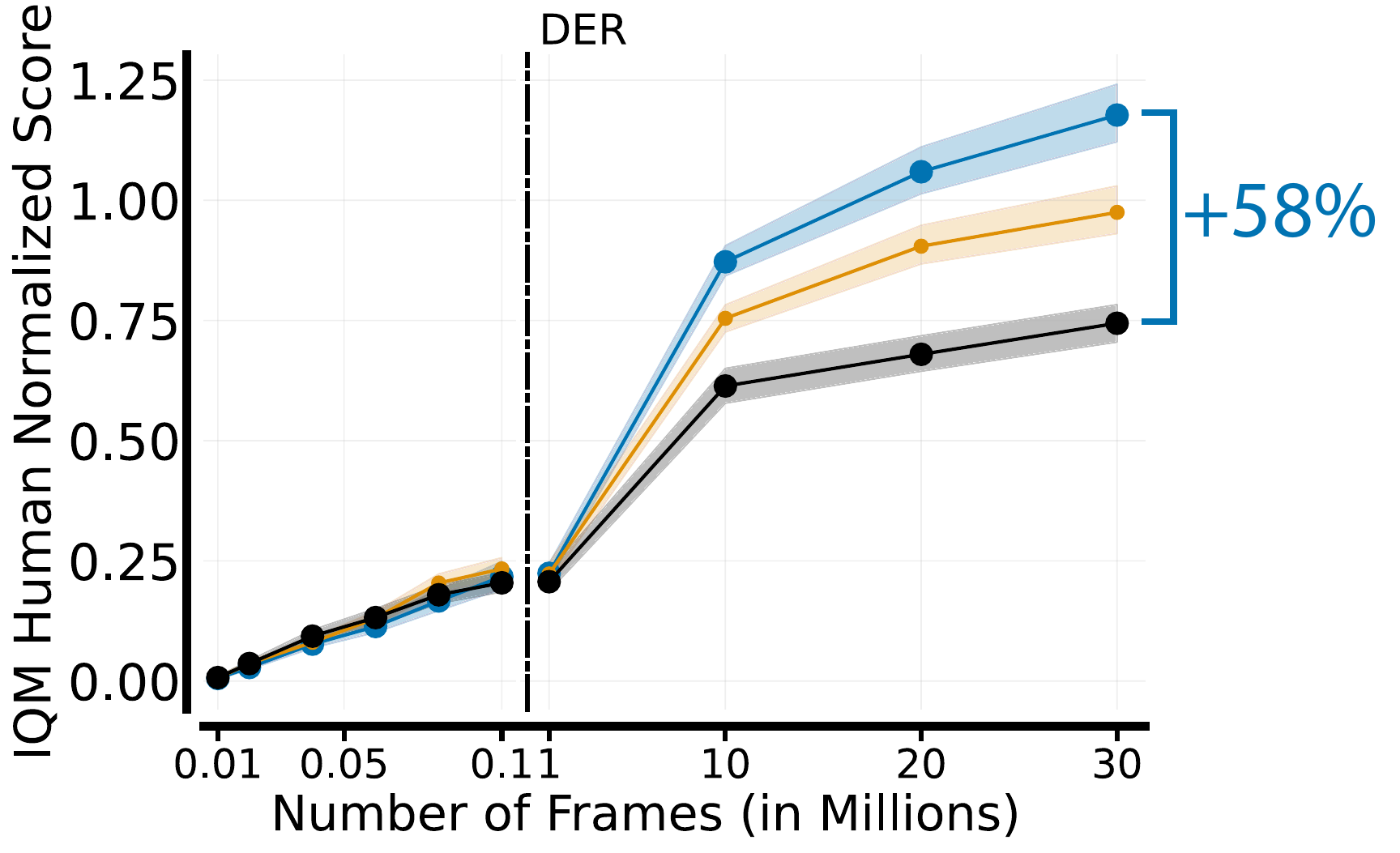}%
    \includegraphics[width=4.6cm]{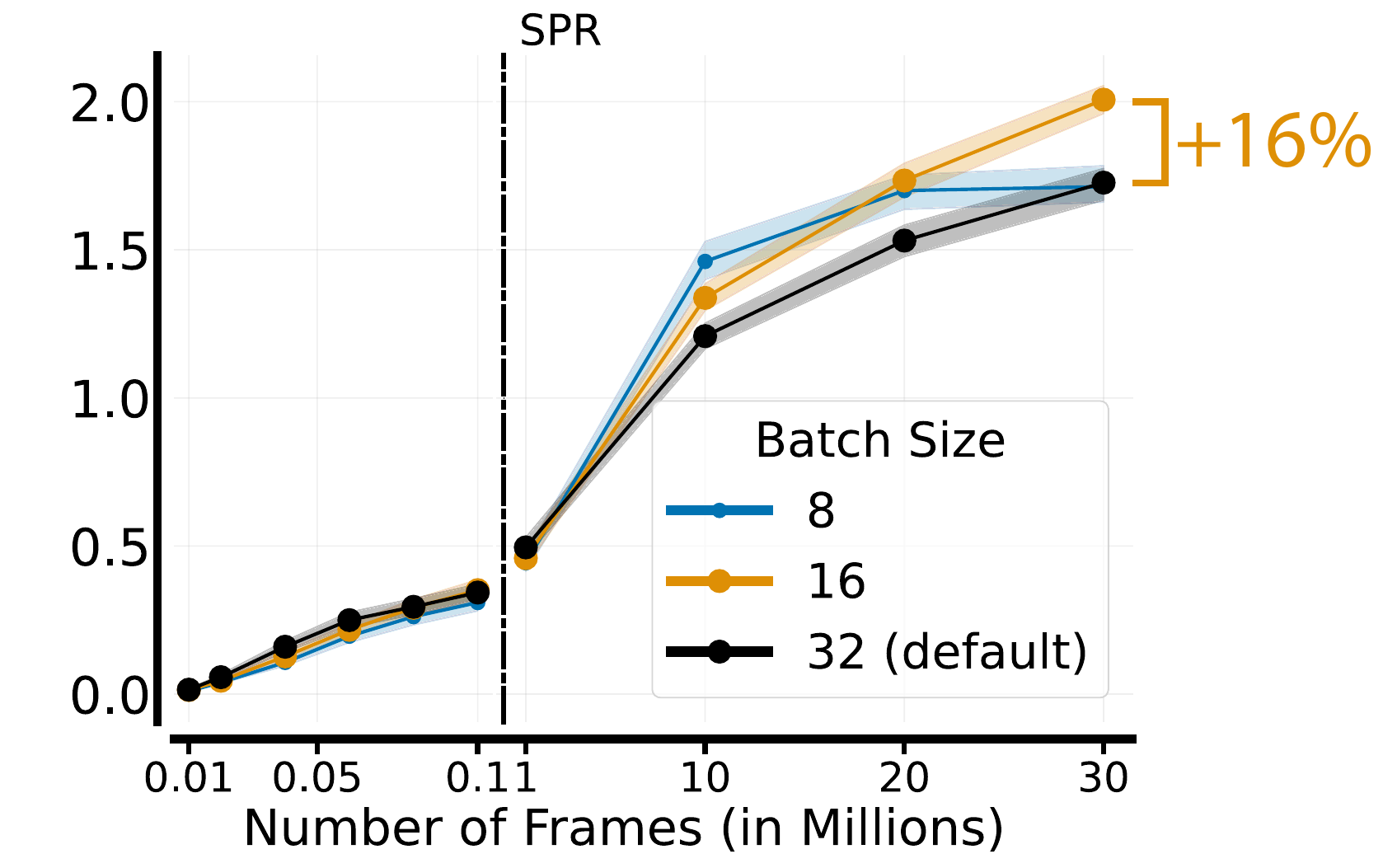}%
   \includegraphics[width=4.6cm]{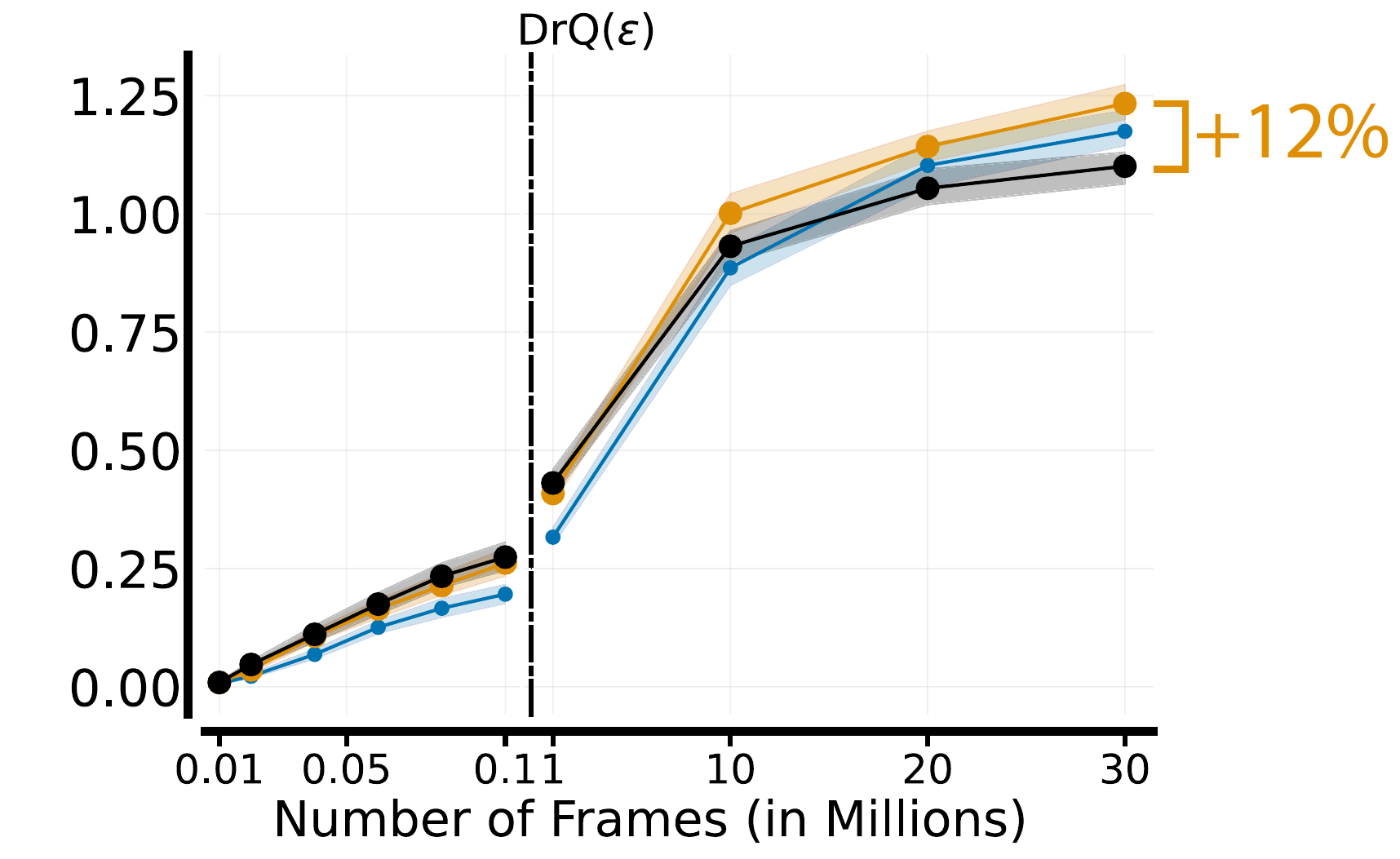}%
    \caption{Measured IQM of human-normalized scores on the 26 100k benchmark games, with varying batch sizes, of DER, SPR, and DrQ($\epsilon$). We evaluate performance at 100k agent steps (or 400k environment frames), and at 30 million environment frames, run with 6 independent seeds for each experiment, and shaded areas display 95\% confidence intervals.}
    \label{fig:aggr100K30M}%
    \vspace{-1em}
\end{figure*}

In \autoref{fig:aggr100K30M} we include results both at the 100k benchmark (left side of plots), and when trained for 30 million frames. Our intent is to evaluate the batch size effect on agents that were optimized for a different training regime. We can see that although there is little difference in 100k, there is a much more pronounced effect when trained for longer. This finding suggests that reduced batch sizes enables continued performance improvements when trained for longer.

\begin{wrapfigure}{r}{0.47\textwidth}
    
    \vspace{-3em}
    \centering
    \includegraphics[width=0.82\linewidth]{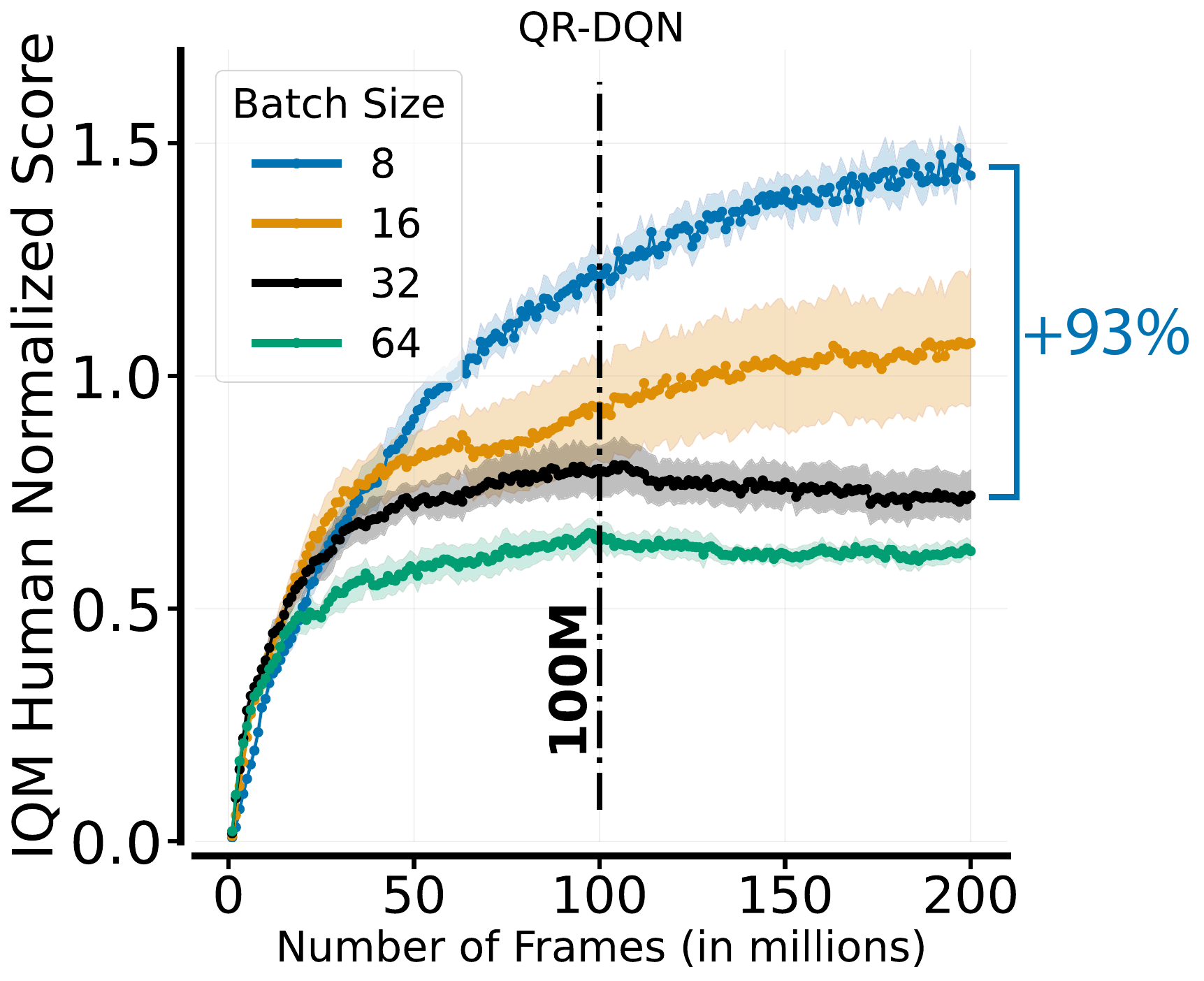}%
\caption{Measuring IQM for human-normalized scores when training for 200 million frames. Results aggregated over 20 games, where each experiment was run with 3 independent seeds and we report 95\% confidence intervals.}%
\label{fig:trainingStabilityIQM}
\vspace{-3em}
\end{wrapfigure}

\subsection{Training Stability} 
\label{sec:trainingStability}
To further investigate whether reduced batch sizes enables continual improvements with longer training, we extend the training of QR-DQN up to the standard 200 million frames. In \autoref{fig:trainingStabilityIQM} we can see that training performance tends to plateau for the higher batch sizes. In contrast, the smaller batch sizes seem to be able to continuously improve their performance.

\subsection{Impact on exploration}
\label{sec:exploration}
\begin{figure*}[!t]
\begin{minipage}{.73\textwidth}
    \includegraphics[width=\linewidth]{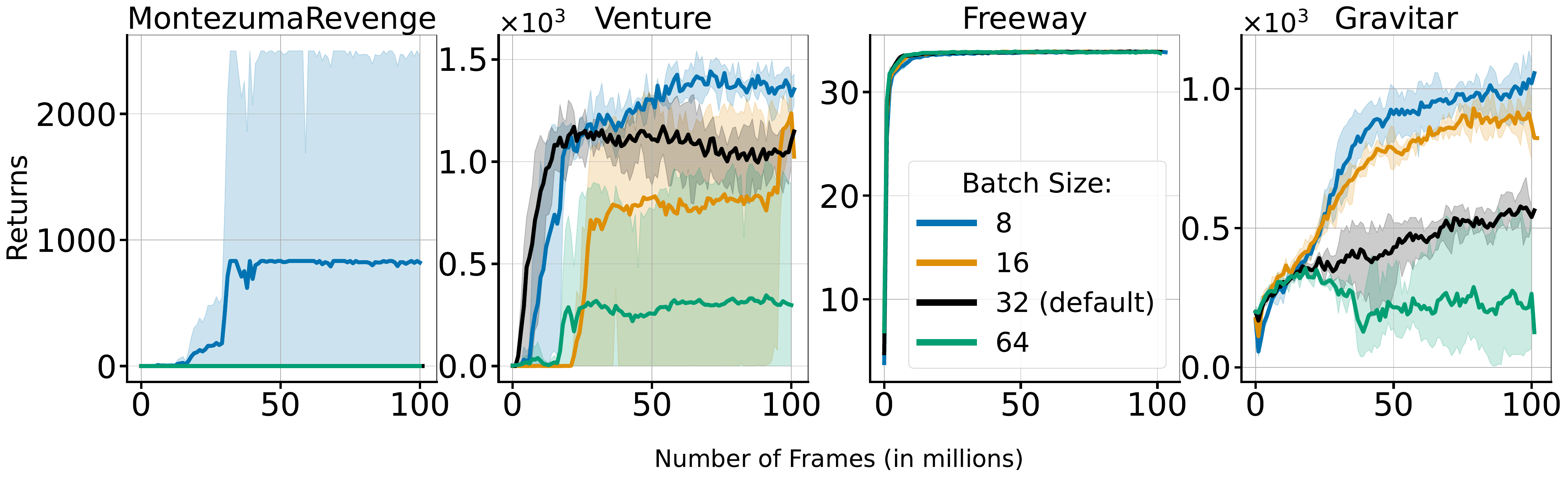}%
    \vline height 90pt depth 0 pt width 1.2 pt
\end{minipage}~~~
\begin{minipage}{.48\textwidth}
    \includegraphics[width=0.55\linewidth]{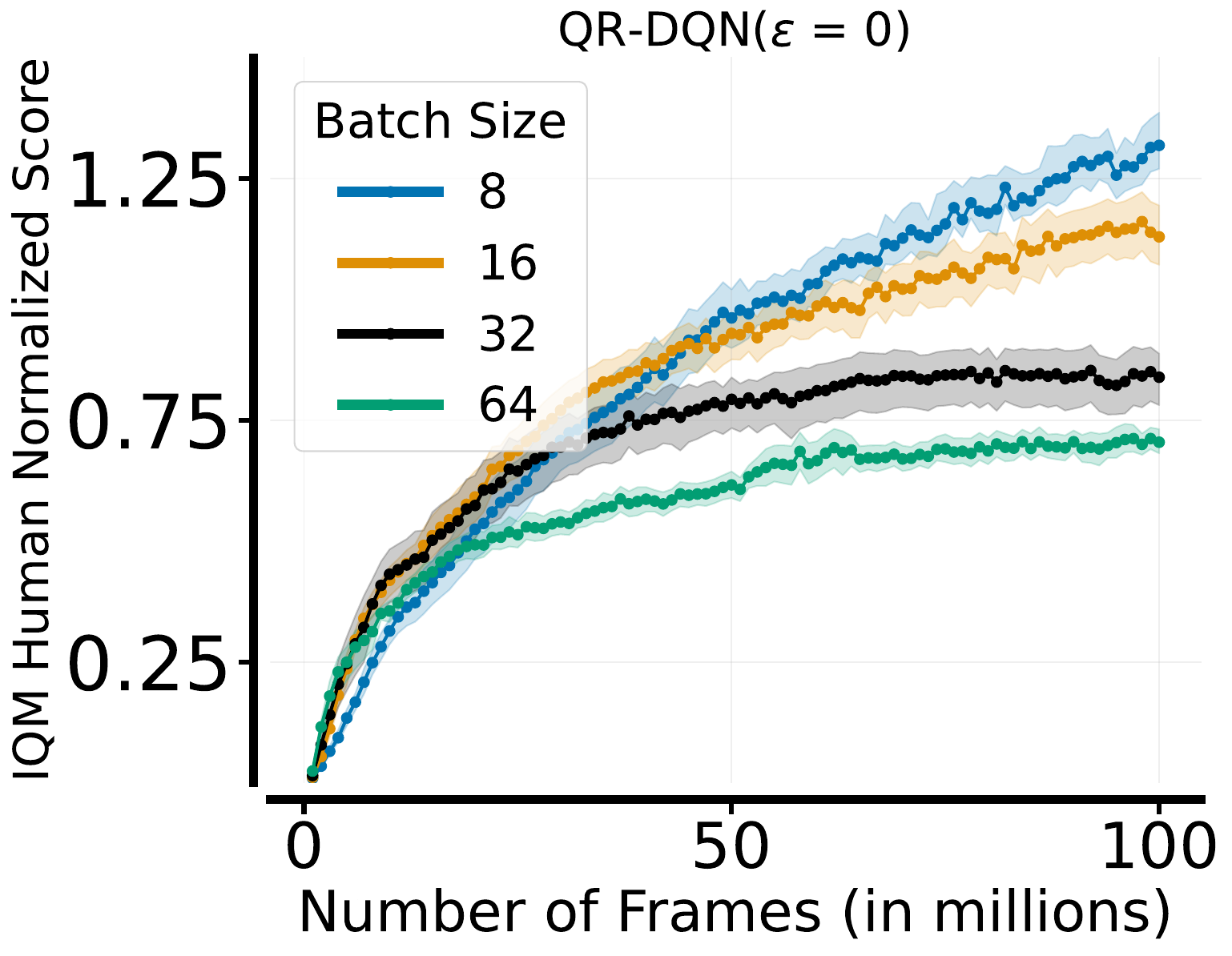}%
    \vfill
\end{minipage}

    \caption{{\bf Left: }Performance of QR-DQN on four hard exploration games with a target $\epsilon$ value of $0.0$, and with varying batch sizes. {\bf Right: } Aggregate IQM of human-normalized scores over 20 games with a target $\epsilon$ value of $0.0$. In all the plots 3 independent seeds were used for each game/batch-size configuration, with shaded areas representing 95\% confidence intervals.}%
    \label{fig:varyingExploration_4games}
    \vspace{-1em}
\end{figure*}

The simplest and most widely used approach for exploration is to select actions randomly with a probability $\epsilon$, as opposed to selecting them greedily from the current $Q_\theta$ estimate. The increased variance resulting from reduced batch sizes (as we will explore in more depth below) may also result in a natural form of exploration. To investigate this, we set the target $\epsilon$ value to $0.0$ for QR-DQN\footnote{Note that we follow the training schedule of \citet{mnih2015humanlevel} where the $\epsilon$ value begins at $1.0$ and is linearly decayed to its target value over the first million environment frames.}. In \autoref{fig:varyingExploration_4games} we compare performance across four known hard exploration games \citep{NIPS2016_afda3322,Taiga2020On} and observe that reduced batch sizes tends to result in improved performance for these games.

Many methods have been proposed to address the exploitation-exploration dilemma, and some techniques emphasize exploration by adding noise directly to the parameter space of agents \citep{fortunato18noisy, plappert2018parameter, Hao_2023, eberhard2023pink}, which inherently adds variance to the learning process. Our analyses show that increasing variance by reducing the batch size may result in similar beneficial exploratory effects, as the mentioned works suggest.



\subsection{Computational impact}
\label{sec:computationalConsequences}
Empirical advances in deep reinforcement learning are generally measured with respect to sample efficiency; that is, the number of environment interactions required before achieving a certain level of performance. It fails to capture computational differences between algorithms. If two algorithms have the same performance with respect to environment interactions, but one takes twice as long to perform each training step, one would clearly opt for the faster of the two. This important distinction, however, is largely overlooked in the standard evaluation methodologies used by the DRL community.

\begin{figure}[!h]
    \centering
    \includegraphics[width=0.248\textwidth]{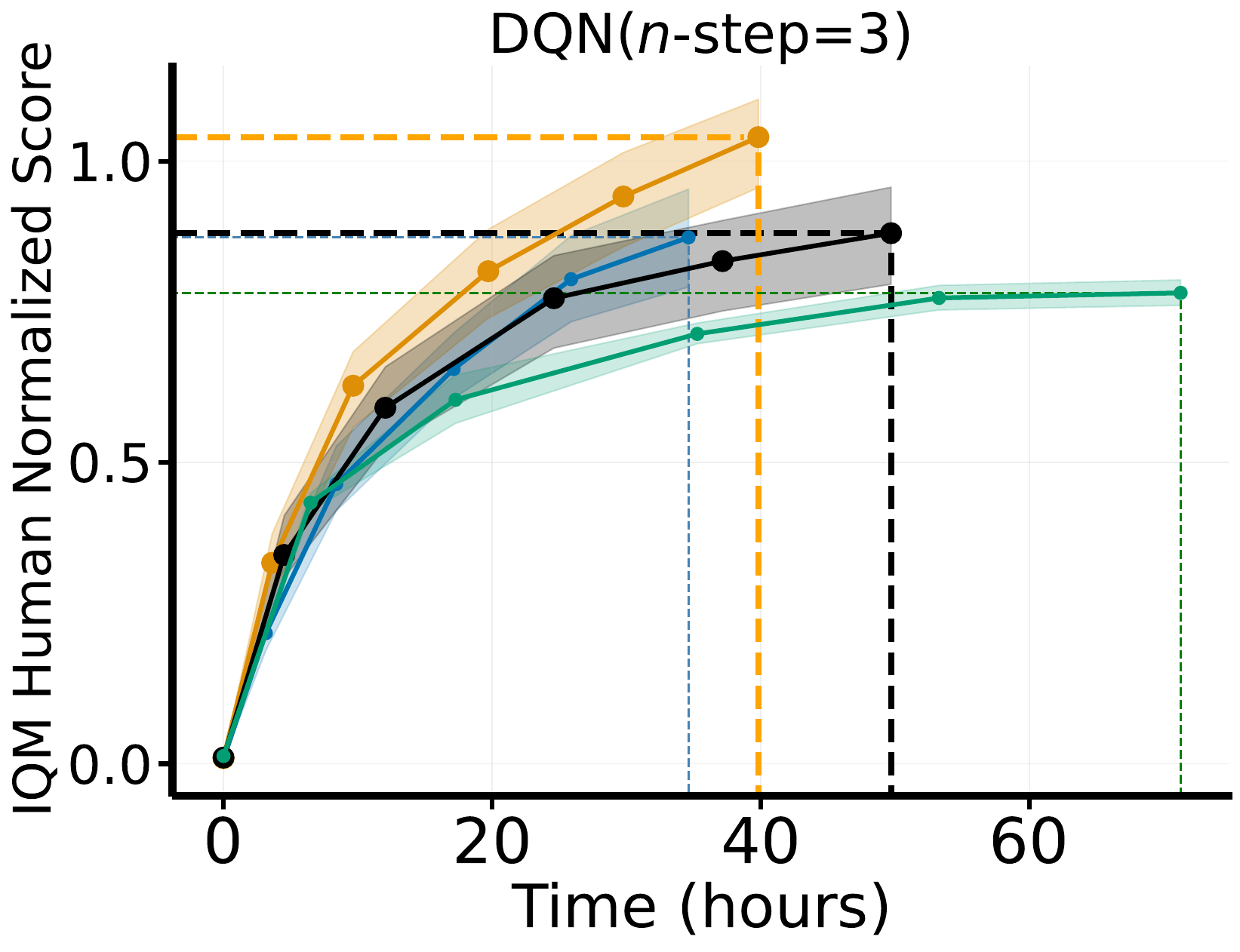}%
    \includegraphics[width=0.248\textwidth]{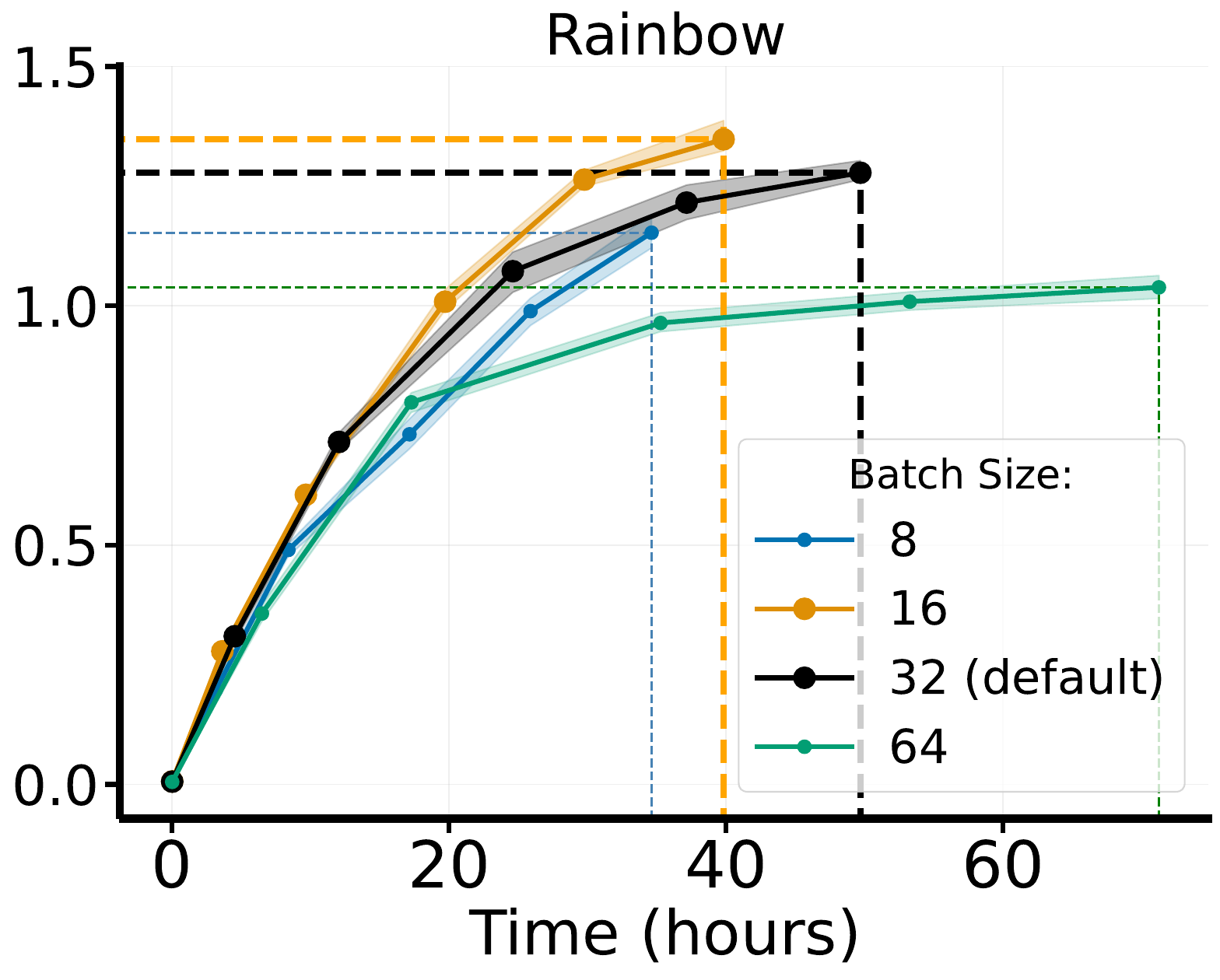}%
   \includegraphics[width=0.248\textwidth]{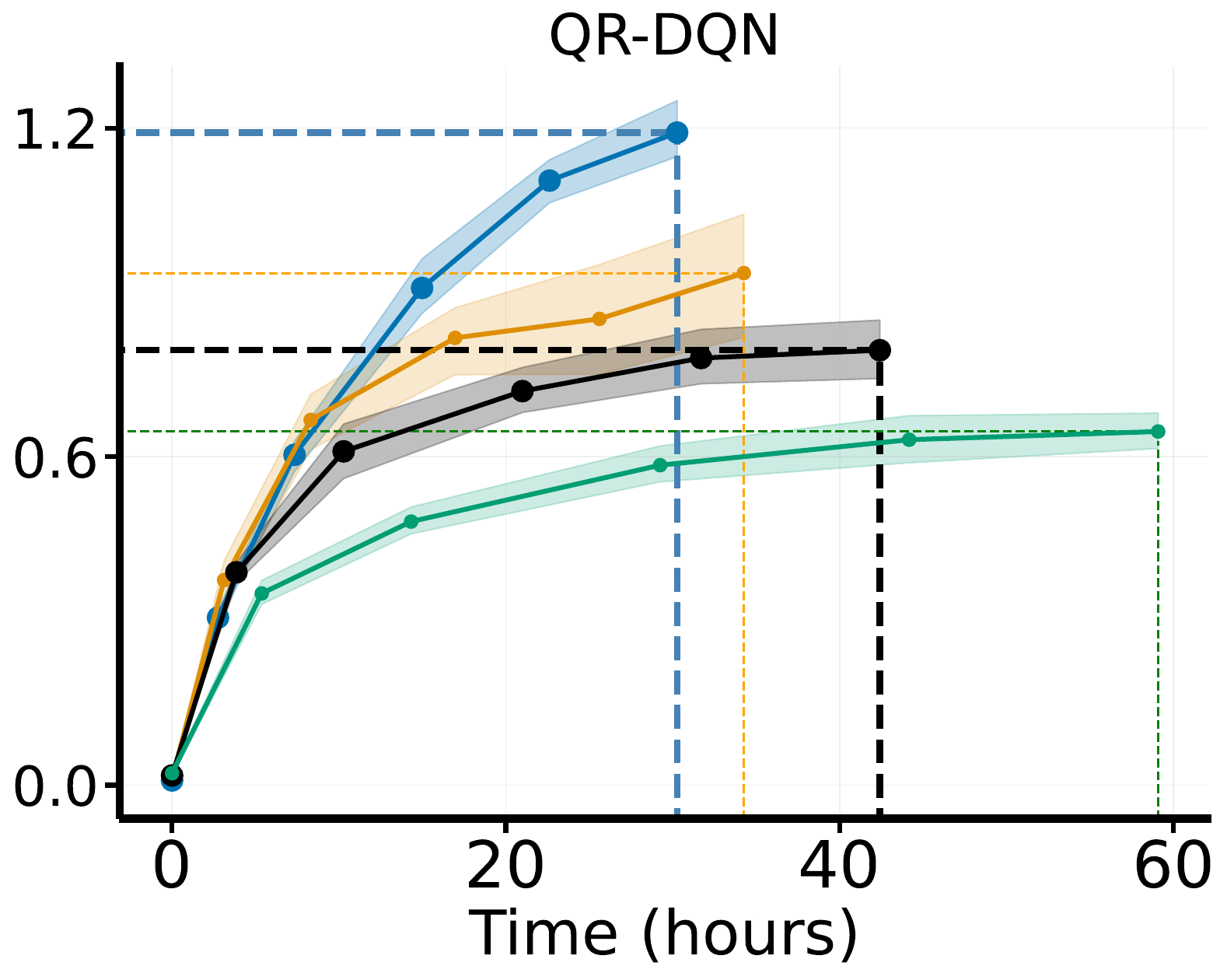}%
    \includegraphics[width=0.248\textwidth]{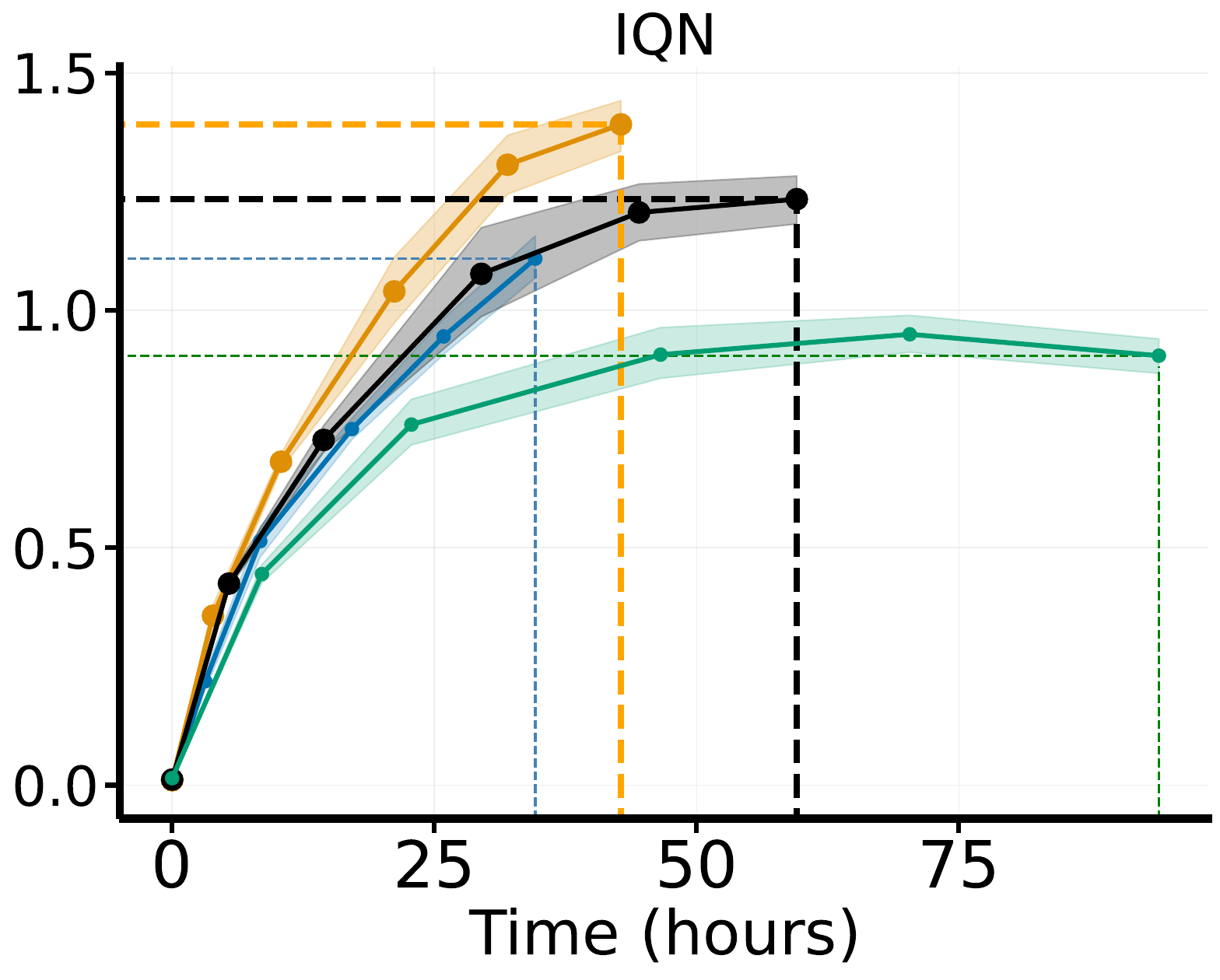}%
    \caption{Measuring wall-time versus IQM of human-normalized scores when varying batch sizes in DQN (with $n$-step set to 3), Rainbow, QR-DQN, and IQN over 20 games. Each experiment had 3 independent runs, and the confidence intervals show 95\% confidence intervals.}%
    \label{fig:runtime_savings_rainbow_qr}%
\end{figure}

We have already demonstrated the performance benefits obtained when reducing batch size, but an additional important consequence is the reduction in computation wall-time. \autoref{fig:runtime_savings_rainbow_qr} demonstrates that not only can we obtain better performance with a reduced batch size, but we can do so at a fraction of the runtime. As a concrete example, when changing the batch size of QR-DQN from the default value of 32 to 8, we achieve both a 50\% performance increase and a 29\% speedup in wall-time. It may seem surprising that smaller batch sizes have a faster runtime, since larger batches presumably make better use of GPU parallelism. However, as pointed out by \citet{Masters2018RevisitingSB}, the speedups may be a result of a smaller memory footprint, enabling better machine throughput.


Considering the unsuitable increase in computational requirements, progress with deep learning demands more compute-efficient training methods. A natural direction is to eliminate algorithmic inefficiencies in the learning process, aiming to reduce time, energy consumption and carbon footprint associated with training these models \citep{bartoldson2023compute, chen2021improving}. \autoref{fig:clock_time_largenets} illustrates the wall-time reduction when using high-capacity neural networks and smaller batch size value. This motivates a fundamental trade-off in the choice of batch size, and the way of how we benchmark deep reinforcement learning algorithms.

\begin{tcolorbox}[leftrule=1.5mm,top=1mm,bottom=0mm]
\textbf{Key observations on reduced batch sizes:}
\begin{itemize}
    \item They generally improve performance, as evaluated across a variety of agents and network architectures.
    \item When trained for longer, the performance gains continue, rather than plateauing.
    \item They seem to have a beneficial effect on exploration.
    \item They result in faster training, as measured by wall-time.
\end{itemize}
\end{tcolorbox}

\section{Understanding the small batch effect}
\label{sec:potential_causes}


Having demonstrated the performance benefits arising from a reduced batch size across a wide range of tasks, in this section we seek to gain some insight into possible causes. We will focus on QR-DQN, as this is the agent where the small batch effect is most pronounced (\autoref{fig:iqmCurveDefaultDQNRainbowQRDQNIQN}). We begin by investigating possible confounding factors for the small batch effect, and then provide analyses on the effect of reduced batch sizes on network dynamics.

\subsection{Relation to other hyperparameters}

\begin{wrapfigure}{r}{0.43\textwidth}
    \vspace{-5.5em}
    \includegraphics[width=\linewidth]{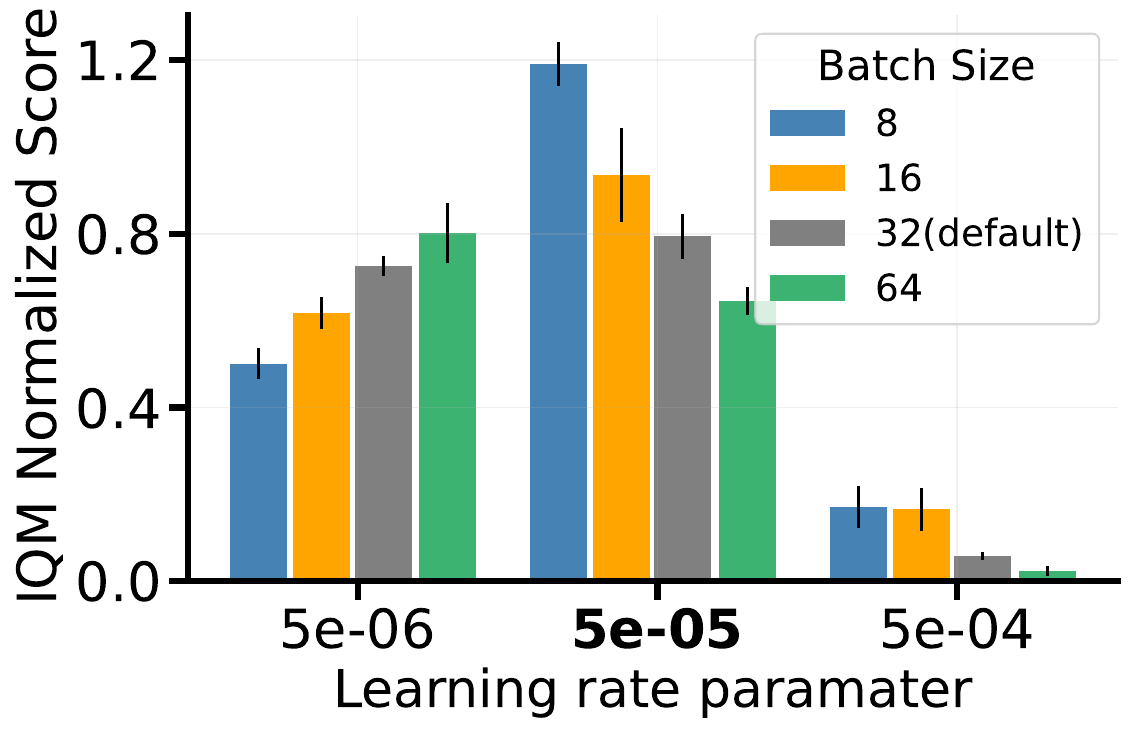}
    \caption{Varying batch sizes for different learning values. Results aggregated IQM of human-normalized scores over 20 games for QR-DQN.}%
    \label{fig:learning_values_eps0}%
    \vspace{-1.5em}
\end{wrapfigure}

\paragraph{Learning rates} It is natural to wonder whether an improved learning rate could produce the same effect as simply reducing the batch size. In \autoref{fig:learning_values_eps0} we explored a variety of different learning rates and observe that, although performance is relatively stable with a batch size of 32, it is unable to reach the performance gains obtained with a batch size of 8 or 16. \autoref{fig:learning_values_eps0} shows that the smaller the learning rate, the larger batch size needs to be, and thus the longer training takes. This result aligns well with  the findings of \citet{WILSON20031429}.

\begin{figure}[!h]
    \centering
   \includegraphics[width=0.75\linewidth]{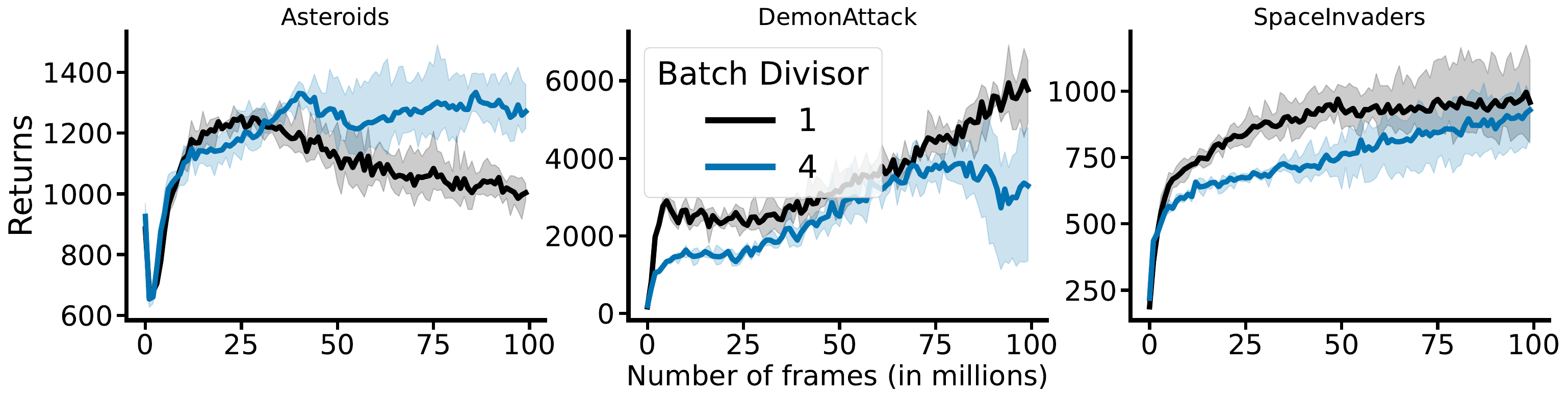}%
    \vline height 75pt depth 0 pt width 1.2 pt
   \includegraphics[width=0.25\linewidth]{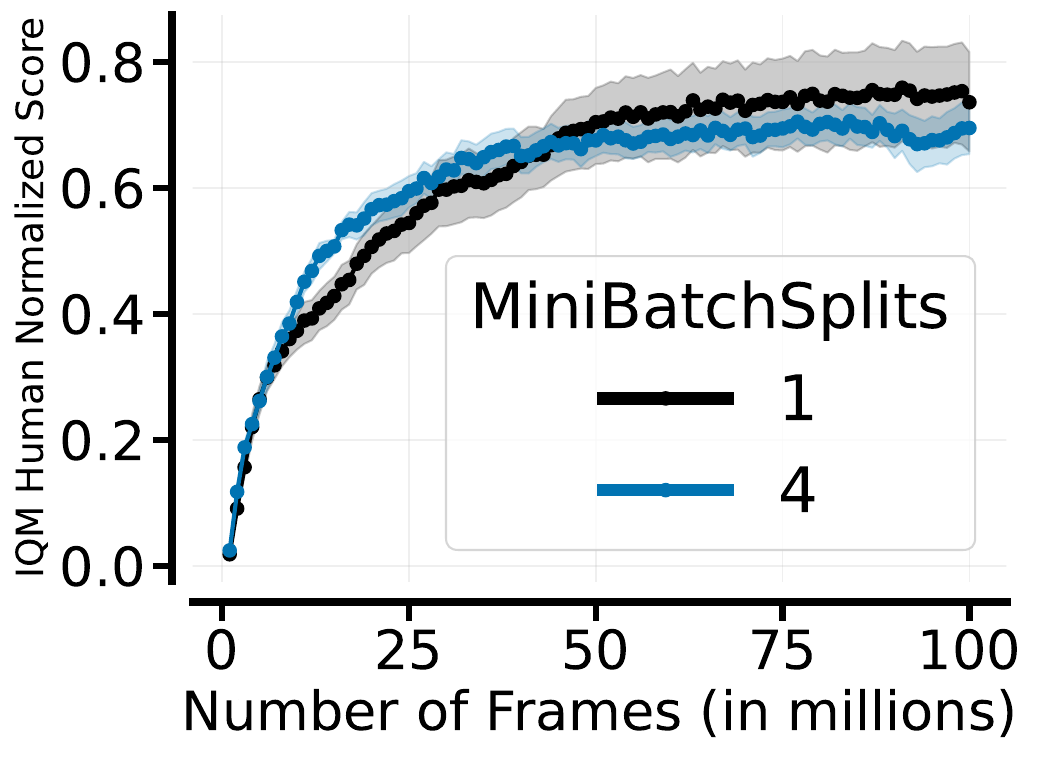}%
    \caption{Varying the number of gradient updates per training step, for a fixed batch size of 32. {\bf Left: }Performance of QR-DQN on three games with different $BatchDivisor$ value. {\bf Right: }Results aggregated IQM of human-normalized scores over 20 games for QR-DQN.}%
    \label{fig:batchDivisor}%
\end{figure}

\paragraph{Second order optimizer effects} All our experiments, like most
modern RL agents, use the Adam optimizer \citep{kingma15adam}, a variant of stochastic gradient
descent (SGD) that adapts its learning rate based on the first- and second-order moments of the
gradients, as estimated from mini-batches used for training. It is thus possible that smaller batch
sizes have a second-order effect on the learning-rate adaptation that benefits agent performance. To
investigate this we evaluated, for each training step, performing multiple gradient updates on subsets
of the original sampled batch; we define the parameter $BatchDivisor$ as the number of gradient updates
and dividing factor (where a value of 1 is the default setting). Thus, for a $BatchDivisor$ of 4, we
would perform 4 gradient updates with subsets of size 8 instead of a single gradient update with a mini-batch of size 32. With an optimizer like SGD this has no effect (as they are mathematically equivalent), but we may see differing performance due to Adam’s adaptive learning rates. \autoref{fig:batchDivisor} demonstrates that, while there are differences, these are not consistent nor significant enough to
explain the performance boost. 

\begin{figure}[!t]
    \centering
   \includegraphics[width=0.69\textwidth]{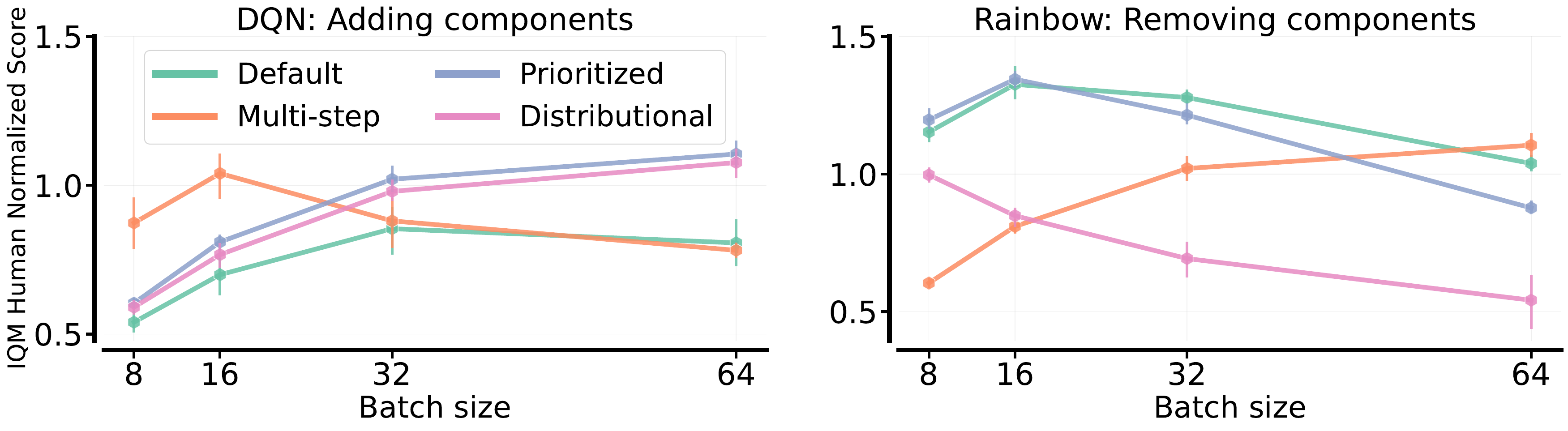}%
   \vline height 80pt depth 0 pt width 1.2 pt
   \includegraphics[width=0.27\textwidth]{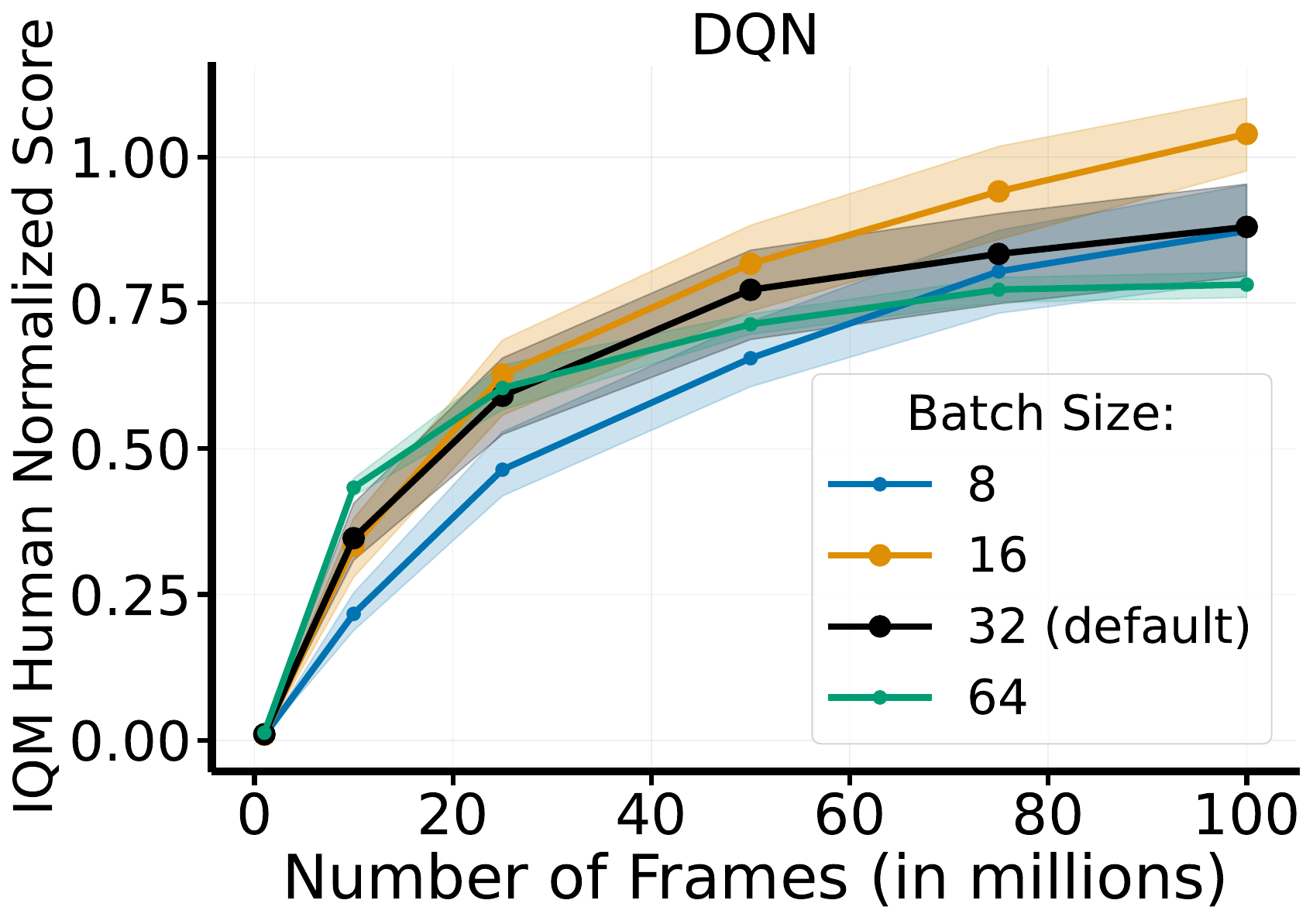}
   \includegraphics[width=\textwidth]{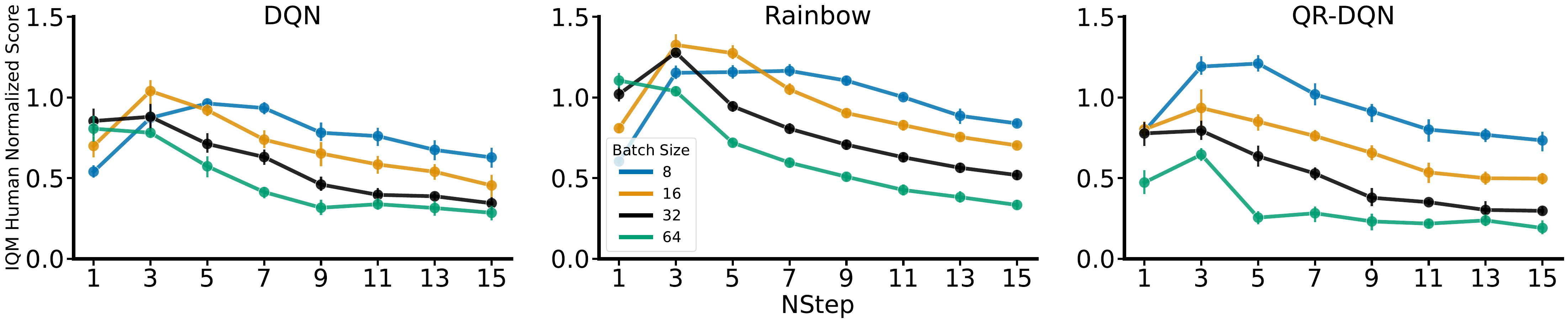}%
    \caption{Measured IQM human normalized scores over 20 games with 3 independent seeds for each configuration, displaying 95\% stratified bootstrap confidence intervals. {\bf Top left:} Adding components to DQN; {\bf Top center:} Removing components from Rainbow. {\bf Top right:} Aggregate DQN performance with $n$-step of 3. {\bf Bottom:} Varying batch sizes and $n$-steps in DQN (left), Rainbow (center), and QR-DQN (right).}
    \label{fig:multiStepAblations}
    \vspace{-1em}
\end{figure}

\paragraph{Relationship with multi-step learning} In \autoref{fig:iqmCurveDefaultDQNRainbowQRDQNIQN} we observed that DQN was the only agent where reducing batch size did not improve performance. Recalling that the Dopamine version of Rainbow used is simply adding three components to the base DQN agent, we follow the analyses of \citet{hessel18rainbow} and \citet{ceron21a}. Specifically, in \autoref{fig:multiStepAblations} (top row) we simultaneously add these components to DQN (top left plot) and remove these components from Rainbow (top center plot). Remarkably, batch size is inversely correlated with performance {\em only when multi-step returns are used}. Given that DQN is the only agent considered here without multi-step learning, this finding explains the anomalous findings in \autoref{fig:iqmCurveDefaultDQNRainbowQRDQNIQN}. Indeed, as the right panel of \autoref{fig:multiStepAblations} (top row) shows, adding multi-step learning to DQN results in improved performance with smaller batch sizes.
To further investigate the relationship between batch size and multi-step returns, in \autoref{fig:multiStepAblations} (bottom row) we evaluate varying both batch sizes and $n$-step values for DQN, Rainbow, and QR-DQN. We can observe that smaller batch sizes suffer less from degrading performance as the $n$-step value is increased.

\begin{tcolorbox}[leftrule=1.5mm,top=1mm,bottom=0mm]
\textbf{Key insights:}
\begin{itemize}
    \item The small batch effect does not seem to be a consequence of a sub-optimal choice of learning rate for the default value of 32.
    \item The small batch effect does not arise due to beneficial interactions with the Adam optimizer.
    \item The small batch effect appears to be more pronounced with multi-step learning.
    \item When increasing the update horizon in multi-step learning, smaller batches produce better results.
\end{itemize}

\end{tcolorbox}

\subsection{Analysis of network optimization dynamics}
 
In this section we will focus on three representative games (Asteroids, DemonAttack, and SpaceInvaders), and include results for more games in the supplemental material. In \autoref{fig:aggregateAnalysesThreeGames} we present the training returns as well as a variety of metrics we collected for our analyses. We will discuss each in more detail below. The first column in this figure displays the training returns for each game, where we can observe the inverse correlation between batch size and performance.

\begin{figure}[!t]
    \centering
   \includegraphics[width=\linewidth]{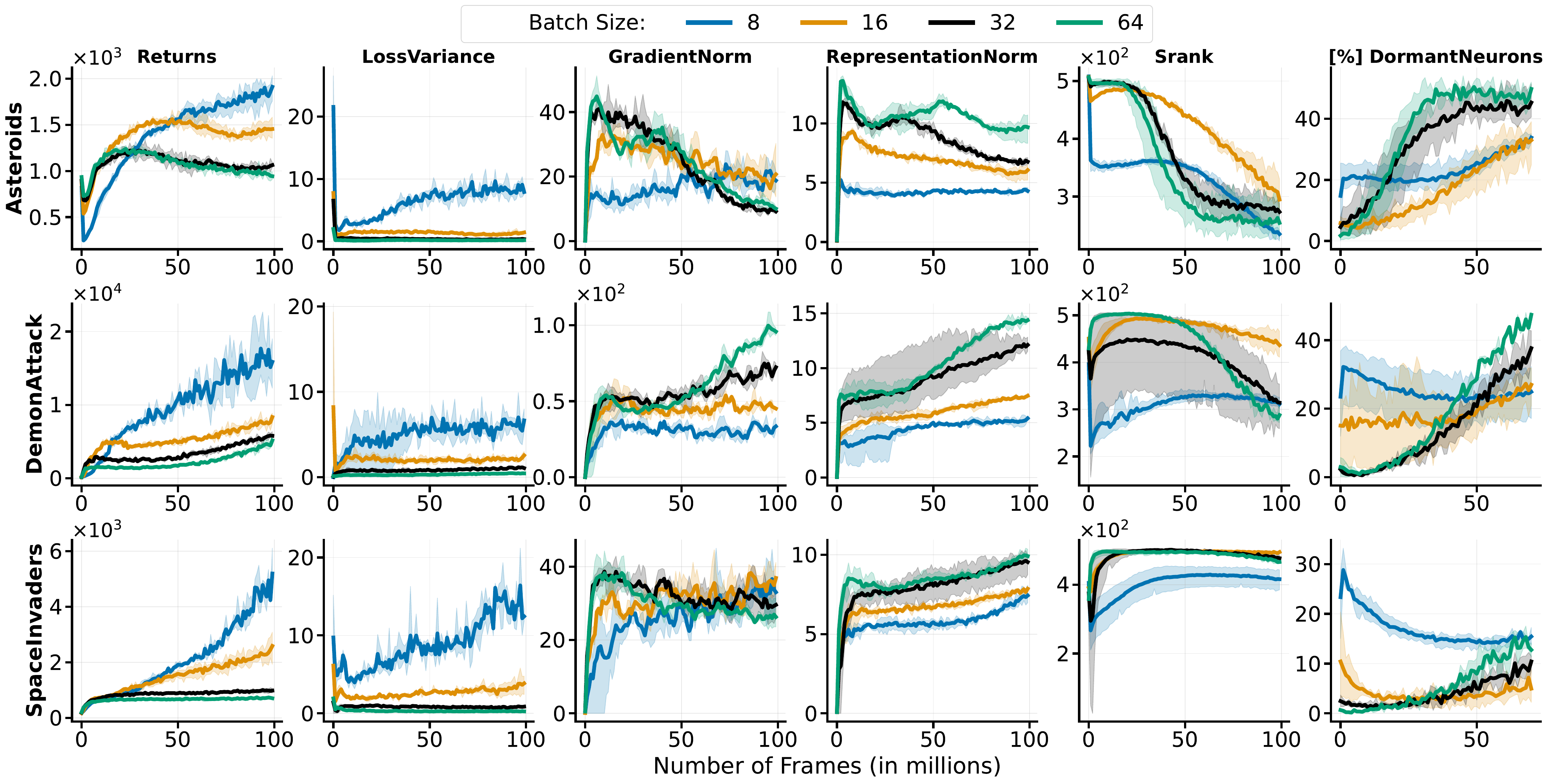}%
    \caption{Empirical analyses for three representative games with varying batch sizes. From left to right: training returns, aggregate loss variance, average gradient norm, average representation norm, $srank$ \citep{kumar2021implicit}, and dormant neurons \citep{sokar23redo}. All results averaged over 3 seeds, shaded areas represent 95\% confidence intervals.}%
    \label{fig:aggregateAnalysesThreeGames}%
    \vspace{-1em}
\end{figure}

\begin{figure}[!h]
    \centering
   \includegraphics[width=0.71\linewidth]{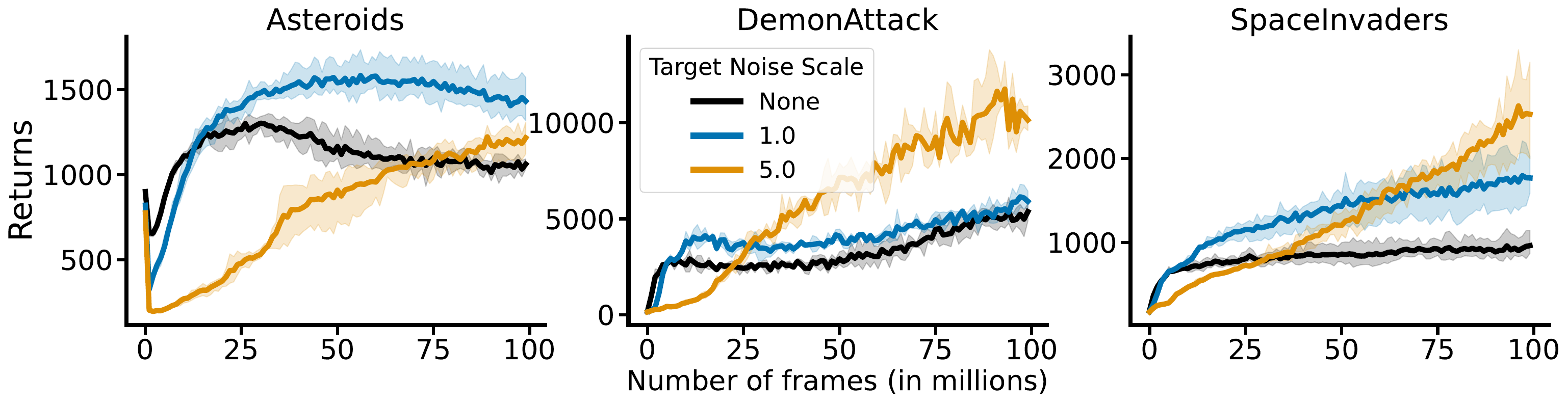}%
   \vline height 75pt depth 0 pt width 1.2 pt
   \includegraphics[width=0.25
   \linewidth]{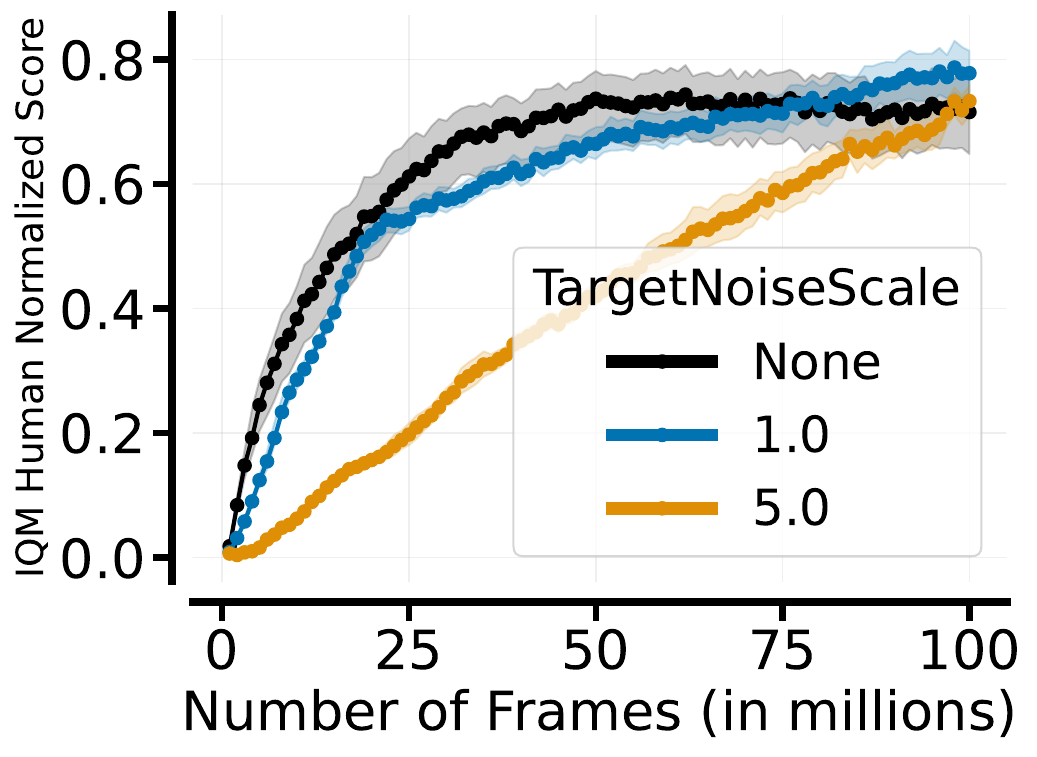}%
    \caption{Adding noise of varying scales to the learning target with the default batch size of 32. {\bf Left: }Performance of QR-DQN on three games with different target noise scale values. {\bf Right: }Results aggregated IQM of human-normalized scores over 20 games for QR-DQN.}%
    \label{fig:targetNoise}%
    \vspace{-1em}
\end{figure}

\paragraph{Variance of updates} Intuition suggests that as we decrease the batch size, we will observe an increase in the variance of our updates as our gradient estimates will be noisier. This is confirmed in the second column of \autoref{fig:aggregateAnalysesThreeGames}, where we see an increased variance with reduced batch size. A natural question is whether directly increasing variance results in improved performance, thereby (partially) explaining the results with reduced batch size. To investigate, we added Gaussian noise (at varying scales) to the learning target $Q_{\bar{\theta}}$ (see \autoref{sec:brackground} for definition). As \autoref{fig:targetNoise} demonstrates, simply adding noise to the target does provide benefits, albeit with some variation across games.

\paragraph{Gradient and representation norms}
\citet{keskar2017on} and \citet{zhao22penalizing} both argue that smaller gradient norms can lead to improved generalization and performance, in part due to less ``sharp'' optimization landscapes. In \autoref{fig:aggregateAnalysesThreeGames} (third column) we can see that batch size is, in fact, correlated with gradient norms, which may be an important factor in the improved performance. In Appendix \ref{sec:average_gradient_norm}, we conducted experiments on a different subset of games, and observed a consistent trend: better performance is achieved with smaller batch sizes and gradient norms.

There have been a number of recent works suggesting RL representations, taken to be the output of the convolutional layers in our networks\footnote{This is a common interpretation used recently, for example, by \citet{castro21mico}, \citet{gogianu21spectral}, and \citet{farebrother2023protovalue}}, yield better agent performance when their norms are smaller. \citet{gogianu21spectral} demonstrated that normalizing representations yields improved agent performance as a result of a change to optimization dynamics;  \citet{kumar2021dr3} further observed that smaller representation norms can help mitigate feature co-adaptation, which can degrade agent performance in the offline setting. As \autoref{fig:aggregateAnalysesThreeGames} (fourth column) shows, the norms of the representations are correlated with batch size, which aligns well with the works just mentioned.

\paragraph{Effect on network expressivity and plasticity} \citet{kumar2021implicit} introduced the notion of the {\em effective rank} of the representation $srank_{\delta}(\phi)$\footnote{$\delta$ is a threshold parameter. We used the same value of $0.01$ as used by \citet{kumar2021implicit}.}, and argued that it is correlated with a network's expressivity: a reduction in effective rank results in an implicit under-parameterization. The authors provide evidence that bootstrapping is the likeliest cause for effective rank collapse (and reduced performance). 
\begin{wrapfigure}{r}{0.5\textwidth}
    \vspace{-1em}
    \centering
    \includegraphics[width=0.5\linewidth]{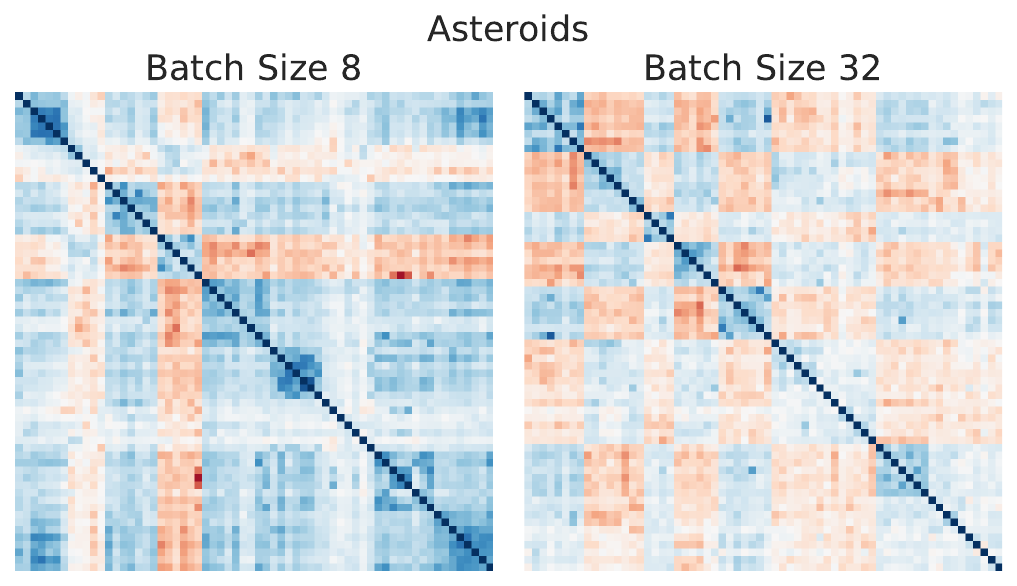}%
    \includegraphics[width=0.5\linewidth]{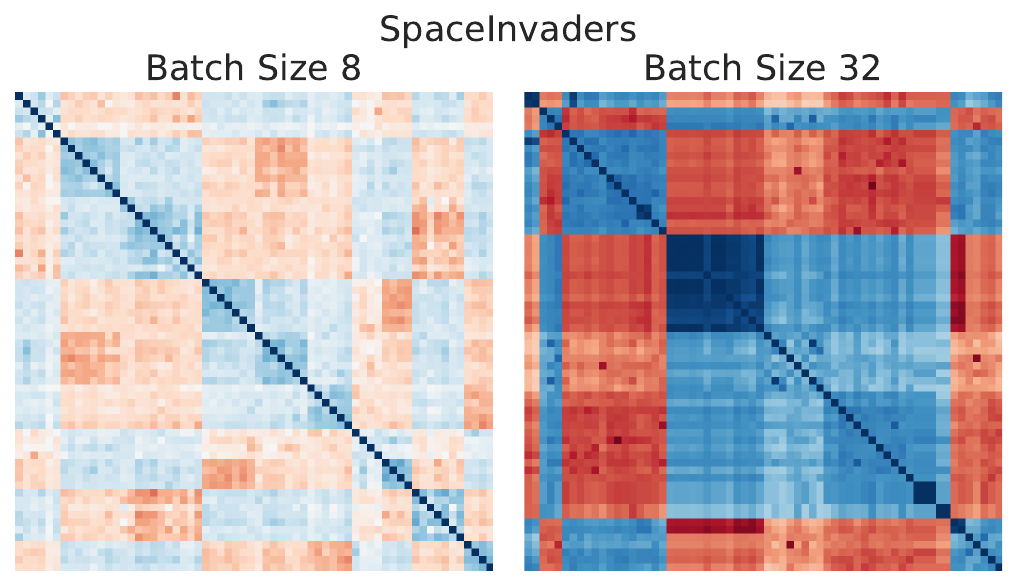}%

\caption[Gradient covariance matrices for Asteroids (\textbf{left}) and SpaceInvaders (\textbf{right}).
In environments where smaller batch size significantly improves
performance, it also induces weaker gradient correlation and less gradient interference.]%
    {Gradient covariance matrices for Asteroids (\textbf{left}) and SpaceInvaders (\textbf{right}).
In environments where smaller batch size significantly improves
performance, it also induces weaker gradient correlation\footref{note1} and less gradient interference. }
    

\label{fig:covariance}
\vspace{-1em}
\end{wrapfigure}
Interestingly, in \autoref{fig:aggregateAnalysesThreeGames} (fifth column) we see that with smaller batch sizes $srank$ collapse occurs earlier in training than with larger batch sizes. Given that there is mounting evidence that deep RL networks tend to overfit during training \citep{dabney21value,nikishin22primacy,sokar23redo}, it is possible that the network is better able to adapt to an earlier rank collapse than to a later one.

To further investigate the effects on network expressivity, we measured the fraction of {\em dormant neurons} (neurons with near-zero activations). \citet{sokar23redo} demonstrated that deep RL agents suffer from an increase in the number of dormant neurons in their network; further, the higher the level of dormant neurons, the worse the performance. In \autoref{fig:aggregateAnalysesThreeGames} (rightmost column) we can see that, although the relationship with batch size is not as clear as with some of the other metrics, smaller batch sizes appear to have a much milder increase in their frequency. Further, there does appear to be a close relationship with the measured $srank$ findings above. \citet{lyle23b} evaluated the covariance structure of the gradients to revisit the network’s loss landscape, and argue that weaker gradient correlation and less gradient interference improve performance. We observe similar results in the gradient covariance heat maps shown in \autoref{fig:covariance} and \autoref{fig:covariance_2}, where gradients appear to be largely colinear\footnote{\label{note1} \textcolor{red}{\textbf{Dark red}} color refers to high negative correlation, and \textcolor{blue}{\textbf{dark blue}} one high positive correlation.} when using larger batch size values.

\begin{tcolorbox}[leftrule=1.5mm,top=1mm,bottom=0mm]
\textbf{Key insights:}
\begin{itemize}
    \item Reduced batch sizes result in increased variance of losses and gradients. This increased variance can have a beneficial effect during training.
    \item Smaller batch sizes result in smaller gradient and representation norms, which tend to result in improved performance.
    \item Smaller batch sizes seem to result in networks that are both more expressive and with greater plasticity.
\end{itemize}
\end{tcolorbox}
\section{Related work}
\label{related_work}

There is a considerable amount of literature on understanding the effect of batch size in supervised learning settings. \citet{keskar2016large} presented quantitative experiments that support the view that large-batch methods tend to converge to sharp minimizers of the training and testing functions, and as has been shown in the optimization community, sharp minima tends to lead to poorer generalization. \citet{Masters2018RevisitingSB} support the previous finding, presenting an empirical study of stochastic gradient descent's performance, and reviewing the underlying theoretical assumptions surrounding smaller batches. They conclude that using smaller batch sizes achieves the best training stability and generalization performance. Additionally, \citet{golmant2018computational} reported that across a
wide range of network architectures and problem domains, increasing the batch size yields no decrease in wall-clock time to convergence for either train or test loss.

Although batch size is central to deep reinforcement learning algorithms, it has not been extensively studied. One of the few results in this space is the work by \cite{stooke18accelerated}, where they argued that larger batch sizes can lead to improved performance when training in distributed settings. Our work finds the opposite effect: {\em smaller} batch sizes tends to improve performance; this suggests that empirical findings may not directly carry over between single-agent and distributed training scenarios. \cite{islam2017reproducibility} and \cite{hilton2022batch} have investigated the role of batch size in on-policy algorithms. The latter demonstrates how to make these algorithms batch size-invariant, aiming to sustain training efficiency at small batch sizes. 

\citet{Lahire2021LargeBE} cast the replay buffer sampling problem as an importance sampling one, allowing it to perform well when using large batch. \citet{fedus2020revisiting} presented a systematic and extensive analysis of experience replay in Q-learning methods, focusing on two fundamental properties: the replay capacity and the ratio of learning updates to experience collected (e.g. the replay ratio). Although their findings are complementary to ours, further investigation into the interplay of batch size and replay ratio is an interesting avenue for future work. 
Finally, there have been a number of recent works investigating network plasticity \citep{schwarzer2023bigger, d'oro2023sampleefficient,sokar23redo, nikishin22primacy}, but all have kept the batch size fixed. 

\citet{wolczyk2021remember} investigate the dynamics of experience replay in online continual learning, and focus on the effect of batch size choice when sampling from a replay buffer. They find that smaller batches are better at preventing forgetting than using larger batches, contrary to the intuitive assumption that it is better to recall more samples from the past to avoid forgetting. Additionally, the authors show that this phenomenon does not disappear under learning rate tuning. Their settings are similar to those used to generate Figure 3 in \citep{sokar23redo}, and suggest that target non-stationarity (e.g. bootstrapping) may have a role to play in explaining the small batch size effect we are observing.

\section{Conclusions}
\label{conclusions}

In online deep RL, the amount of data sampled during each training step is crucial to an agent's learning effectiveness. Common intuition would lead one to believe that larger batches yield better estimates of the data distribution and yield computational savings due to data parallelism on GPUs. Our findings here suggest the opposite: the batch size parameter generally alters the agent's learning curves in surprising ways, and reducing the batch size below its standard value is often beneficial.

From a practical perspective, our experimental results make it clear that the effect of batch size on performance is substantially more complex than in supervised learning. Beyond the obvious performance and wall-time gains we observe, changing the batch size appears to have knock-on effects on exploration as well as asymptotic behaviour. Figure \ref{fig:learning_values_eps0} hints at a complex relationship between learning rate and batch size, suggesting the potential usefulness of ``scaling laws'' for adjusting these parameters appropriately.

Conversely, our results also highlight a number of theoretically-unexplained effects in deep reinforcement learning. For example, one would naturally expect that decreasing the batch size should increase variance, and eventually affect prediction accuracy. That its effect on performance, both transient and asymptotic, should so critically depend on the degree to which bootstrapping occurs (as in $n$-step returns; Figure \ref{fig:multiStepAblations}), suggests that gradient-based temporal-difference learning algorithms need a fundamentally different analysis from supervised learning methods.

\paragraph{Future Work}


Our focus in this paper has been on value-based online methods. This raises the question of whether our findings carry over to actor-critic methods, and different training scenarios such as offline RL \citep{levine2020offline} and distributed training \citep{stooke18accelerated}. While similar findings are likely for actor-critic methods, the dynamics are sufficiently different in offline RL and in distributed training that it would likely require a different investigative and analytical approach. It is also an interesting direction to explore adaptive schemes that dynamically varies the batch size during training. Our experiments used a constant batch size, so further research is needed to determine whether it is advantageous to reduce the batch size over time in practice, as well as how quickly it should be reduced.

Our work has broader implications than just the choice of the batch size hyper-parameter. For instance, our findings on the impact of variance on performance suggest a promising avenue for new algorithmic innovations via the explicit injection of variance. Most exploration algorithms are designed for tabular settings and then adapted for deep networks; our results in section \ref{sec:exploration} suggest there may be opportunities for exploratory algorithms designed specifically for use with neural networks. We hope our analyses can prove useful for further advances in the development and understanding of deep networks for reinforcement learning.

    
    

\newpage
\paragraph{Acknowledgements.} Many thanks to Georg Ostrovski and Gopeshh Subbaraj for their feedback on an earlier draft of this paper. We also acknowledge Max Schwarzer, Adrien Ali Taiga, Rishabh Agarwal and Jesse Farebrother for useful discussions, as well as the rest of the DeepMind Montreal team for their feedback on this work. The authors would also like to thank the anonymous reviewers for useful feedback on this paper. Last but not least, we would also like to thank the Python community \citep{van1995python, 4160250} for developing tools that enabled this work, including NumPy \citep{harris2020array}, Matplotlib \citep{hunter2007matplotlib} and JAX \citep{bradbury2018jax}.

\paragraph{Broader impact} Although the work presented here 
is mostly of an academic nature, it aids in the development of more capable autonomous agents. While our
contributions do not directly contribute to any negative
societal impacts, we urge the community to consider
these when building on our research

\bibliographystyle{plainnat}
\bibliography{references}

\newpage
\newpage
\appendix
\onecolumn

\section{Code availability}
\label{sec:code}
Our experiments were built on open source code, mostly from the Dopamine repository. The root directory for these is \href{https://github.com/google/dopamine/tree/master/dopamine/}{https://github.com/google/dopamine/tree/master/dopamine/}, and we specify the subdirectories below (with clickable links):
\begin{itemize}
    \item DQN, Rainbow, QR-DQN and IQN agents from \href{https://github.com/google/dopamine/tree/master/dopamine/jax/agents}{/jax/agents/} 
    \item Atari-100k agents from \href{https://github.com/google/dopamine/tree/master/dopamine/labs/atari_100k}{/labs/atari-100k/}
    \item Batch size from  \href{https://github.com/google/dopamine/blob/master/dopamine/jax/agents/quantile/configs/quantile.gin#L36}{/jax/agents/quantile/configs/quantile.gin} \textbf{(line 36)}
    \item Exploration $\epsilon=0$ from  \href{https://github.com/google/dopamine/blob/master/dopamine/jax/agents/quantile/configs/quantile.gin#L16}{/jax/agents/quantile/configs/quantile.gin} \textbf{(line 16)}
    \item Resnet from \href{https://github.com/google/dopamine/blob/master/dopamine/labs/offline_rl/jax/networks.py#L108}{/labs/offline-rl/jax/networks.py} \textbf{(line 108)}
    \item Dormant neurons metric from \href{https://github.com/google/dopamine/tree/master/dopamine/labs/redo}{/labs/redo/}
\end{itemize}

For the srank metric experiments we used code from: \\ \href{https://github.com/google-research/google-research/blob/master/generalization_representations_rl_aistats22/coherence/coherence_compute.py}{https://github.com/google-research/google-research/blob/master/\\     generalization\_representations\_rl\_aistats22/coherence/coherence\_compute.py}

\section{Atari 2600 games used}
\label{sec:atari_2600}

Most of our experiments were run with 20 games from the ALE suite \citep{Bellemare_2013}, as suggested by \citet{fedus2020revisiting}. However, for the Atari 100k agents (\autoref{sec:lowDataRegime}), we used the standard set of 26 games \citep{Kaiser2020Model} to be consistent with the benchmark. Finally, we also ran some experiments with the full set of 60 games. The specific games are detailed below.

\textbf{20 game subset:} AirRaid, Asterix, Asteroids, Bowling, Breakout, DemonAttack, Freeway, Gravitar, Jamesbond, MontezumaRevenge, MsPacman, Pong, PrivateEye, Qbert, Seaquest, SpaceInvaders, Venture,         WizardOfWor, YarsRevenge, Zaxxon.

\textbf{26 game subset:} Alien, Amidar, Assault, Asterix, BankHeist, BattleZone, Boxing, Breakout, ChopperCommand, CrazyClimber, DemonAttack, Freeway, Frostbite, Gopher, Hero, Jamesbond, Kangaroo, Krull, KungFuMaster, MsPacman, Pong, PrivateEye, Qbert, RoadRunner, Seaquest, UpNDown.

\textbf{60 game set:} The 26 games above in addition to: AirRaid, Asteroids, Atlantis, BeamRider, Berzerk, Bowling, Carnival, Centipede, DoubleDunk, ElevatorAction, Enduro, FishingDerby, Gravitar, IceHockey, JourneyEscape, MontezumaRevenge, NameThisGame, Phoenix, Pitfall, Pooyan, Riverraid, Robotank, Skiing, Solaris, SpaceInvaders, StarGunner, Tennis, TimePilot, Tutankham, Venture, VideoPinball, WizardOfWor, YarsRevenge, Zaxxon.

\section{Wall-time versus IQM of human-normalized}
\label{sec:gradient_covarianc}

\begin{figure}[!h]
    \centering
    \vfill
    \includegraphics[width=6cm]{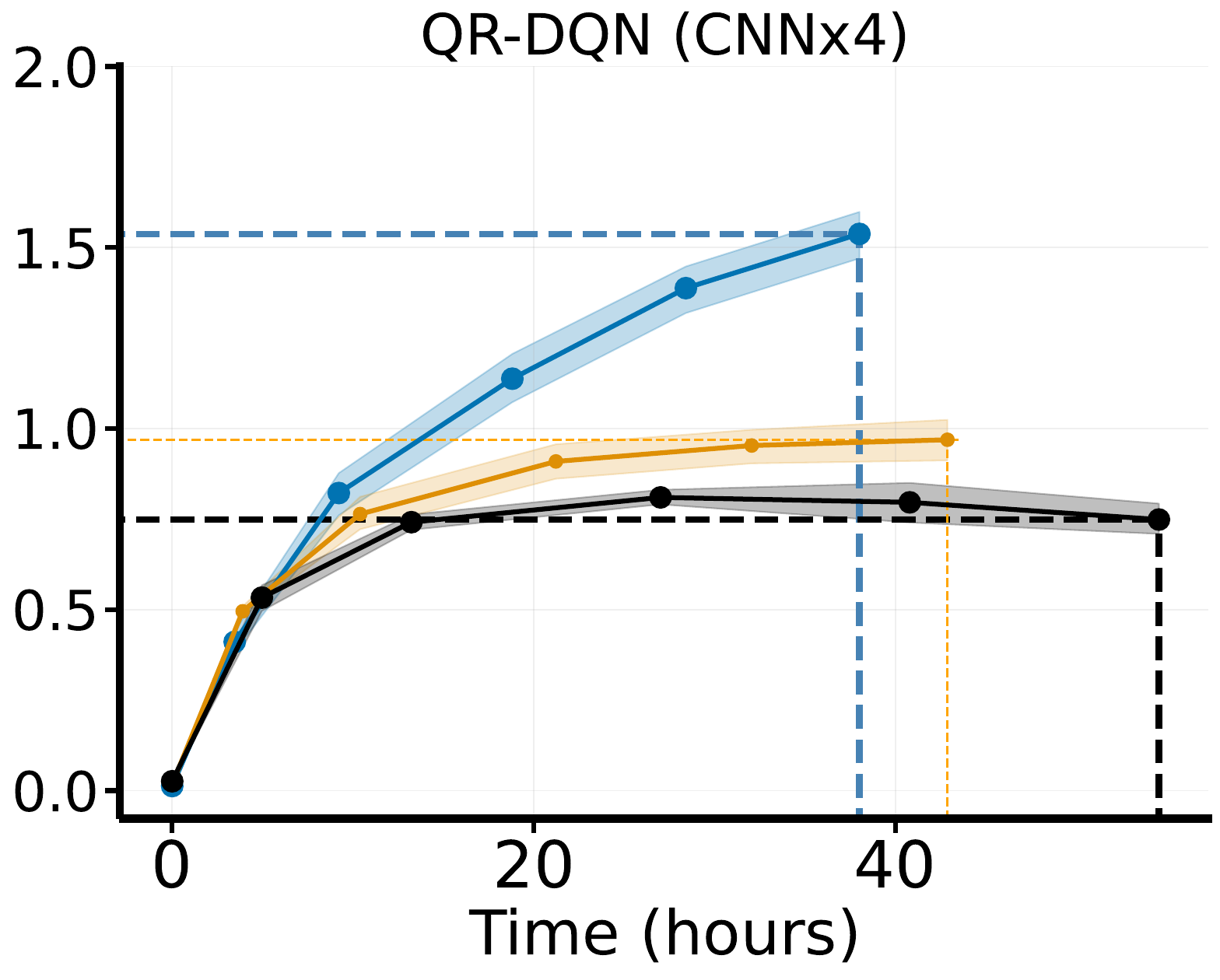}%
    \includegraphics[width=6cm]{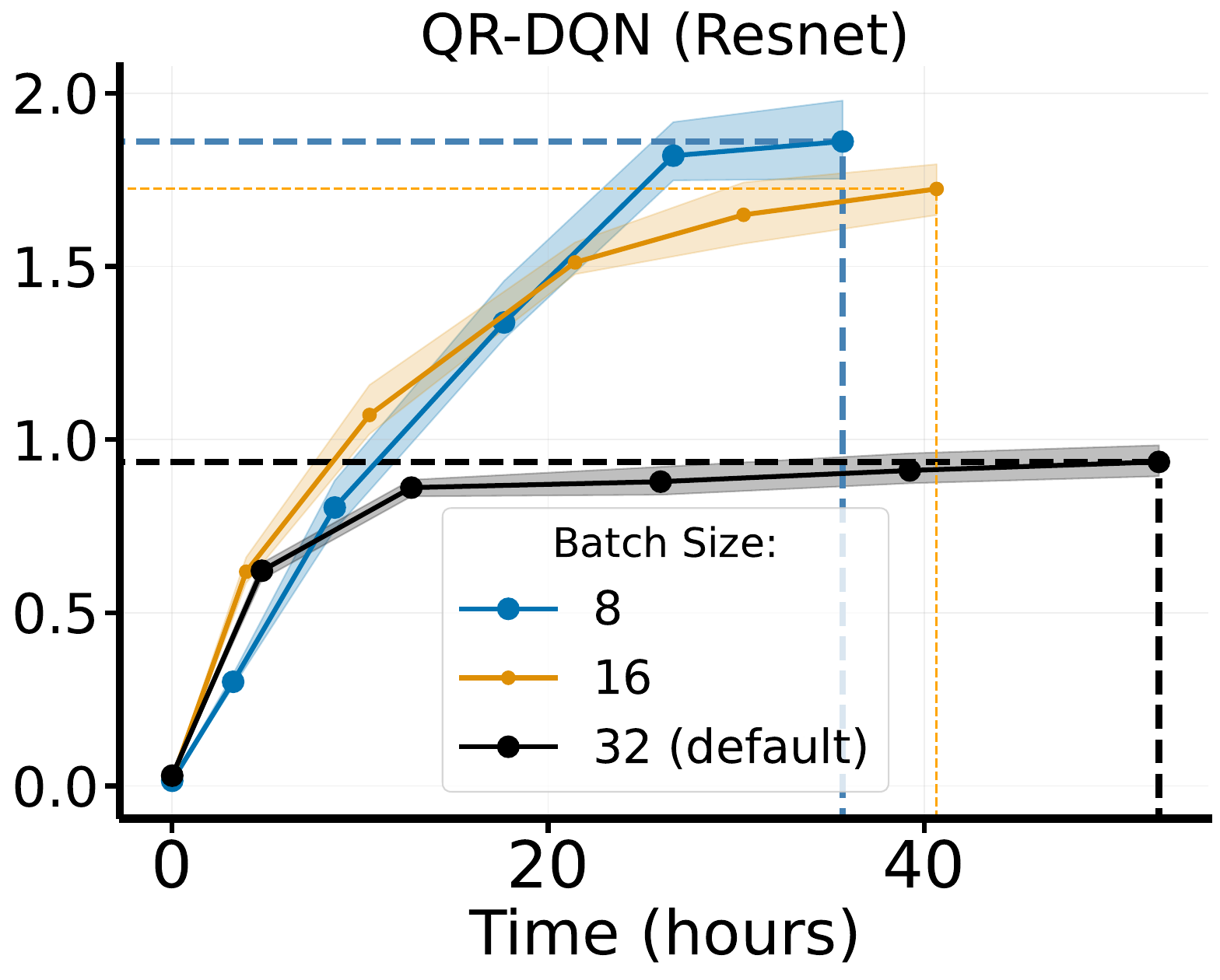}%
    \caption{Measuring wall-time versus IQM of human-normalized scores when varying batch sizes and neural network architectures over 20 games in QR-DQN. Each experiment had 3 independent runs, and the confidence intervals show 95\% confidence intervals.
}
\label{fig:clock_time_largenets}
\end{figure}



\newpage
\section{Average gradient norm}
\label{sec:average_gradient_norm}

\begin{figure}[!h]
    \centering

    \includegraphics[width=1\textwidth]{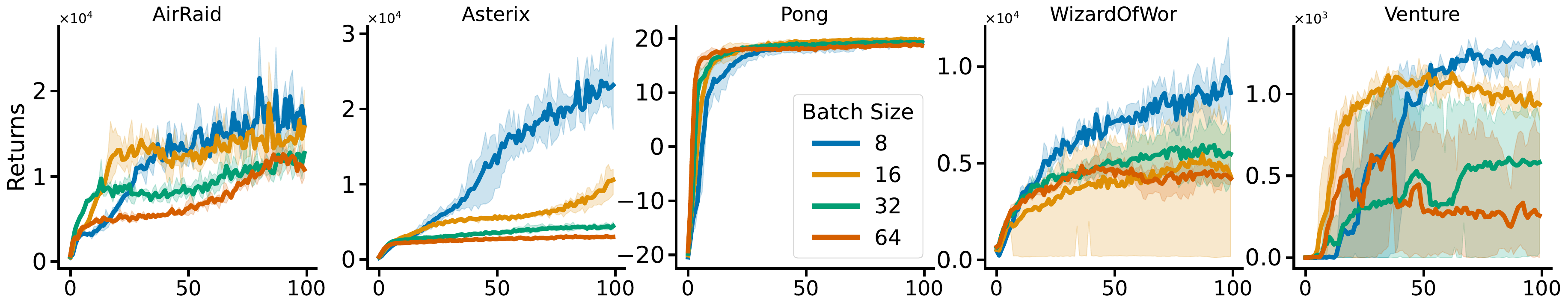}
    \includegraphics[width=1\textwidth]{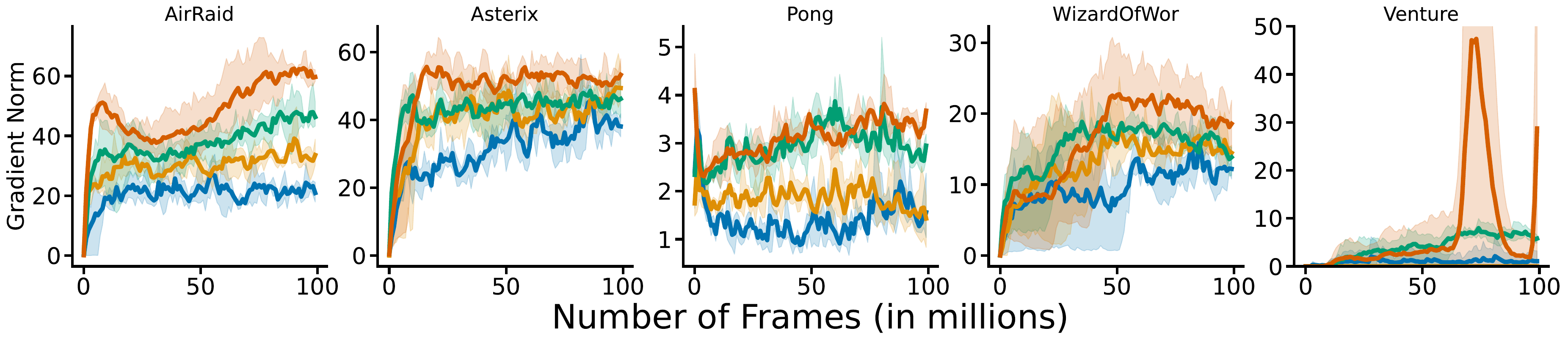}
    \caption{Empirical analyses for 5 representative games with varying batch sizes. \textbf{Top:} training returns, \textbf{Bottom: } average gradient norm. Results averaged over 3 seeds, shaded areas represent 95\% confidence intervals.}%
    \label{fig:average_gradient_norm}%
\end{figure}

\section{Gradient covariance}
\label{sec:gradient_covarianc}

\begin{figure}[!h]
    \centering
    \vfill
    \includegraphics[width=4.6cm]{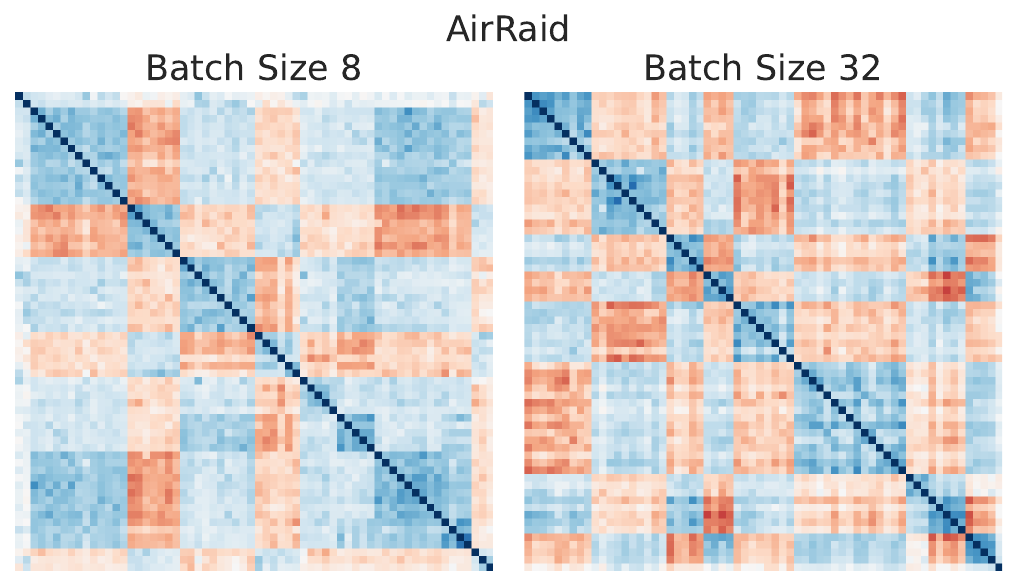}%
    \includegraphics[width=4.6cm]{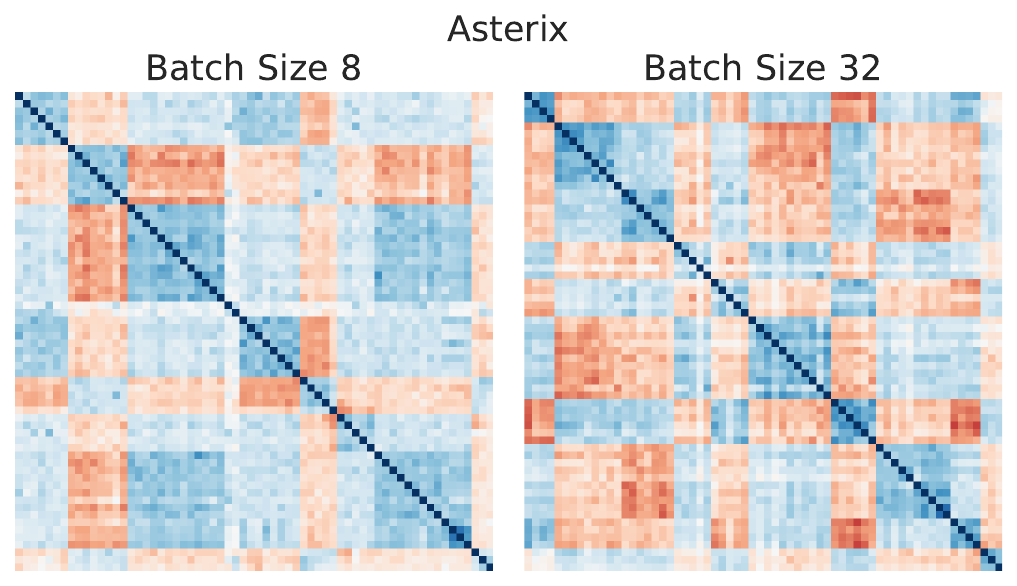}%
   \includegraphics[width=4.6cm]{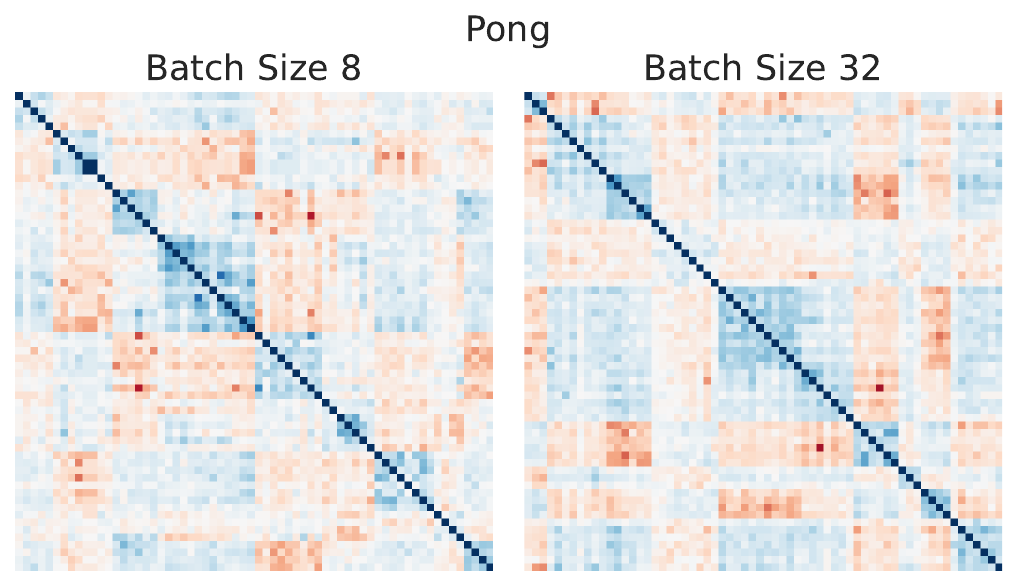}
   
   \includegraphics[width=4.6cm]{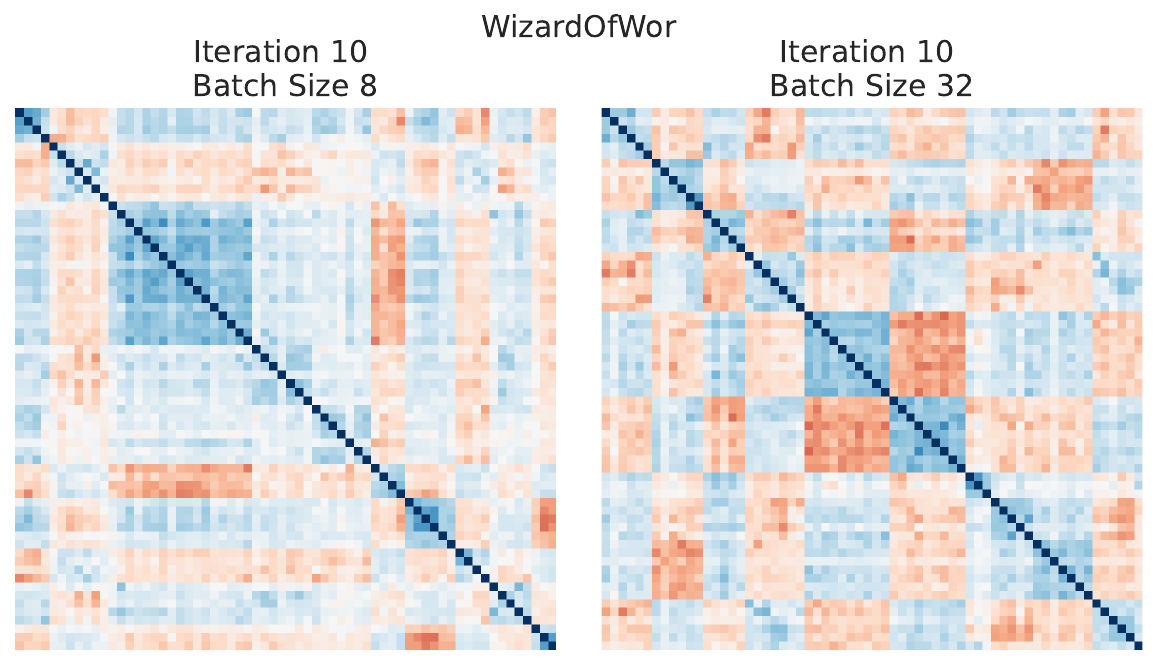}%
    \includegraphics[width=4.6cm]{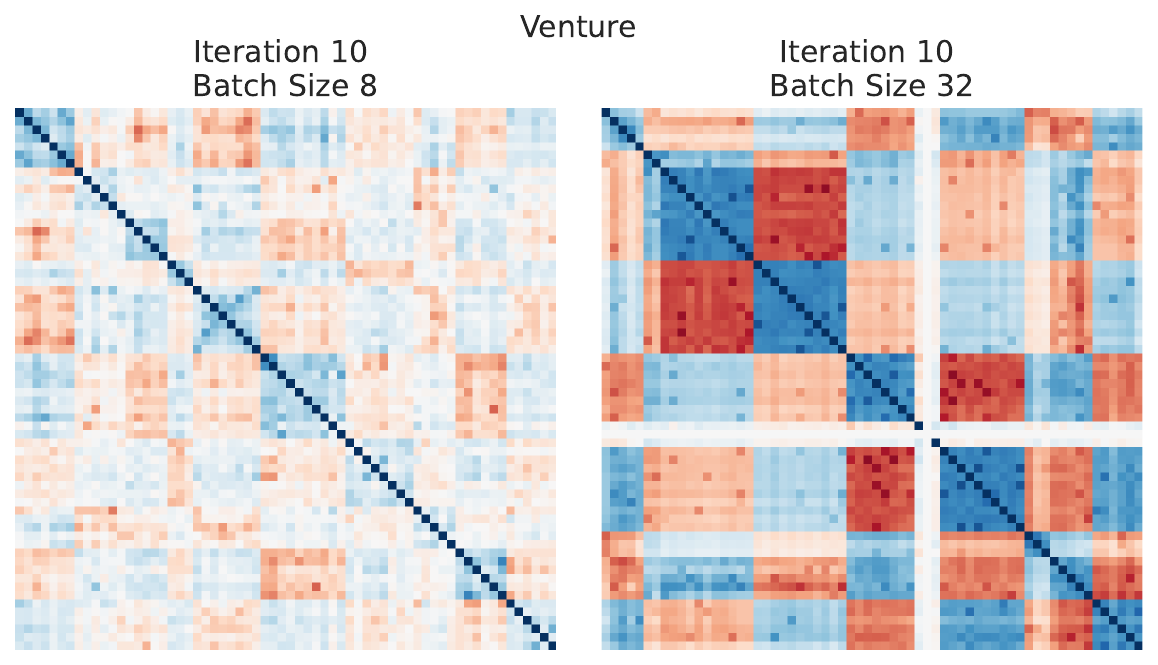}%
   \includegraphics[width=4.6cm]{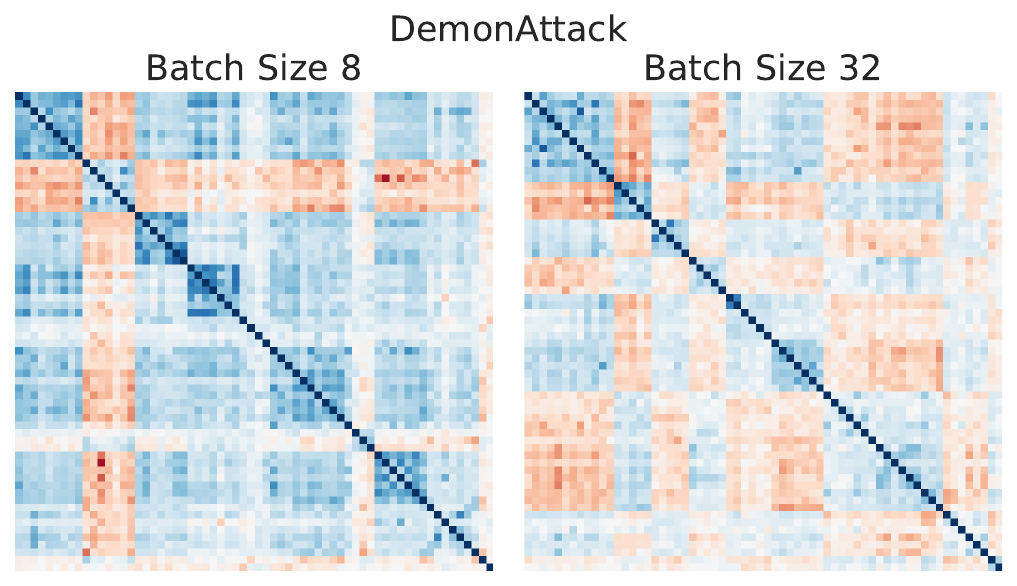}%
   
    \caption{Gradient covariance plots for 6 representative games, which highlight the role of the gradient structure with varying batch sizes. We find that smaller batch size significantly improves performance and induces less gradient interference and weaker gradient correlation.
}
\label{fig:covariance_2}
\end{figure}
 
\newpage
\section{Second order optimizer effects}
\label{sec:appendixMiniMiniBatch}


\begin{figure}[!h]
    \centering
    \includegraphics[width=0.835\textwidth]{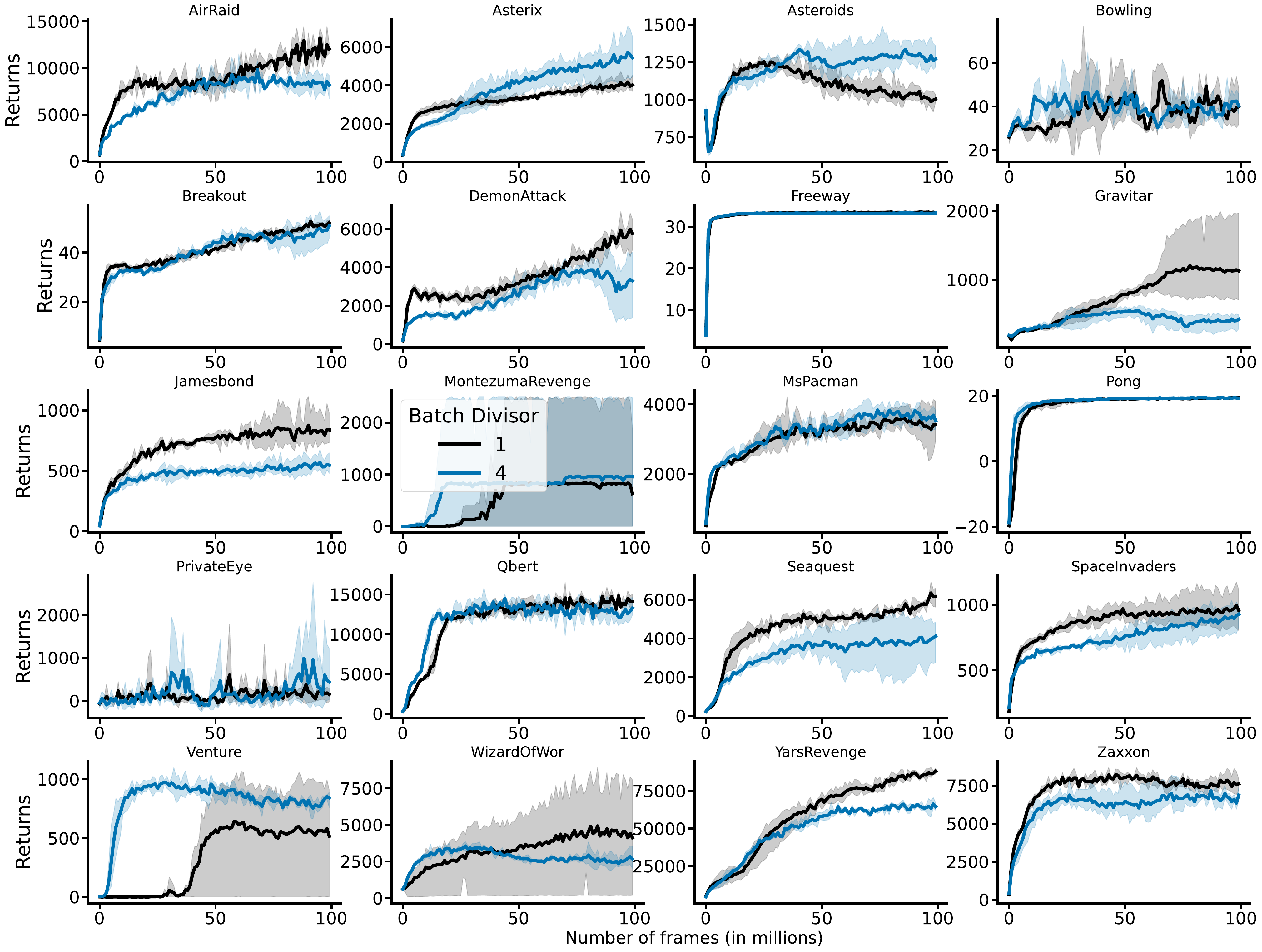}%
    \caption{Evaluating multiple gradient updates per training step on QR-DQN, training curves for all games. Results averaged over 3 seeds, shaded areas represent 95\% confidence intervals.}
    \label{fig:batchDivisorSplitGames}
\end{figure}

\section{Variance of updates.}
\label{sec:appendixVariancePlots}

\begin{figure}[!h]
    \centering
    \includegraphics[width=0.835\textwidth]{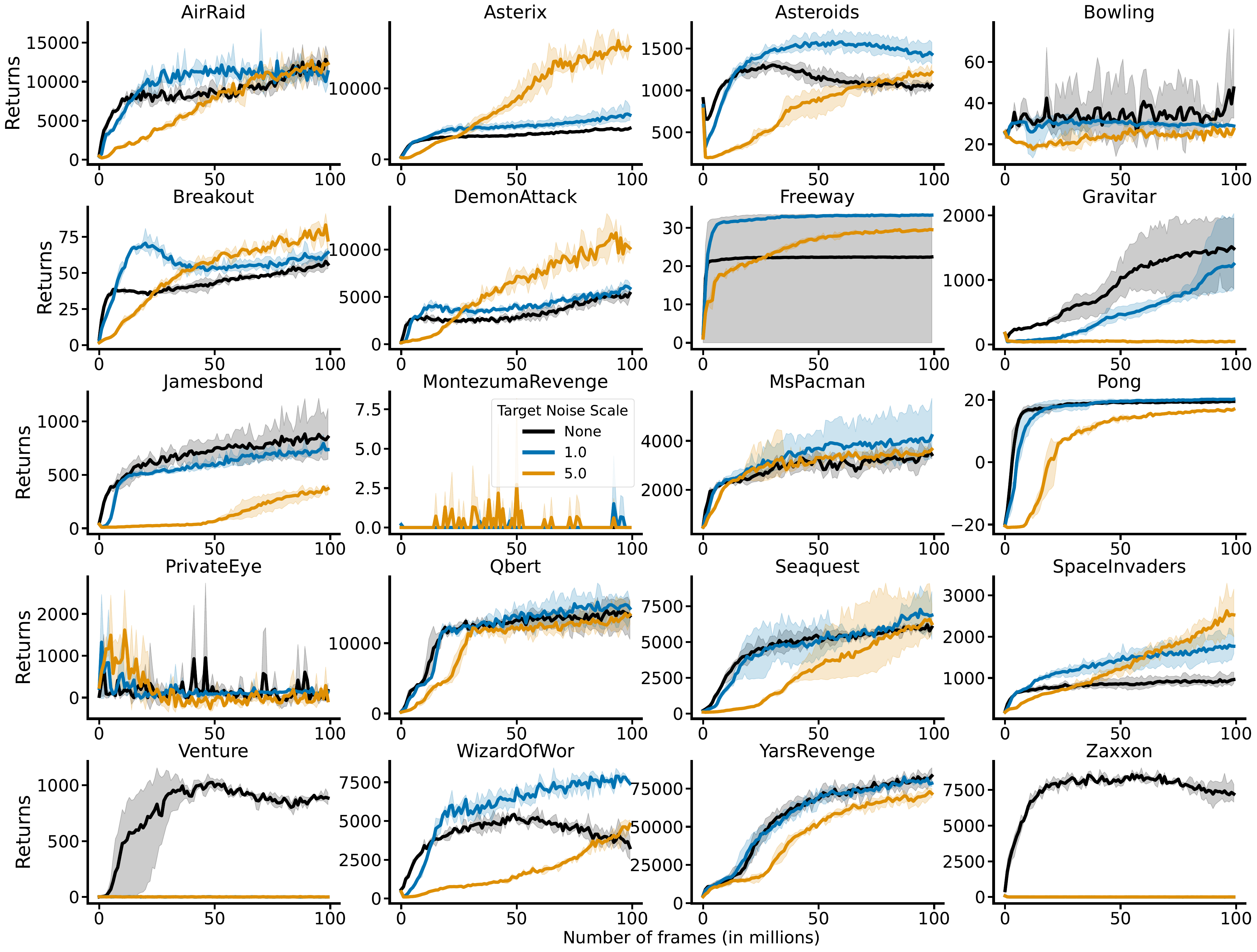}
    \caption{Evaluating the effect of adding target noise to QR-DQN, learning curves for all games. Results averaged over 3 seeds, shaded areas represent 95\% confidence intervals.}
    \label{fig:targetNoiseAllGames}
\end{figure}

\newpage
\section{Results on the full ALE suite}
\label{sec:all_games}
We additionally provide complete results for all games using QR-DQN agent in Figure \ref{fig:all_games}.

\begin{figure}[!h]
    \centering
    \includegraphics[width=0.9\textwidth]{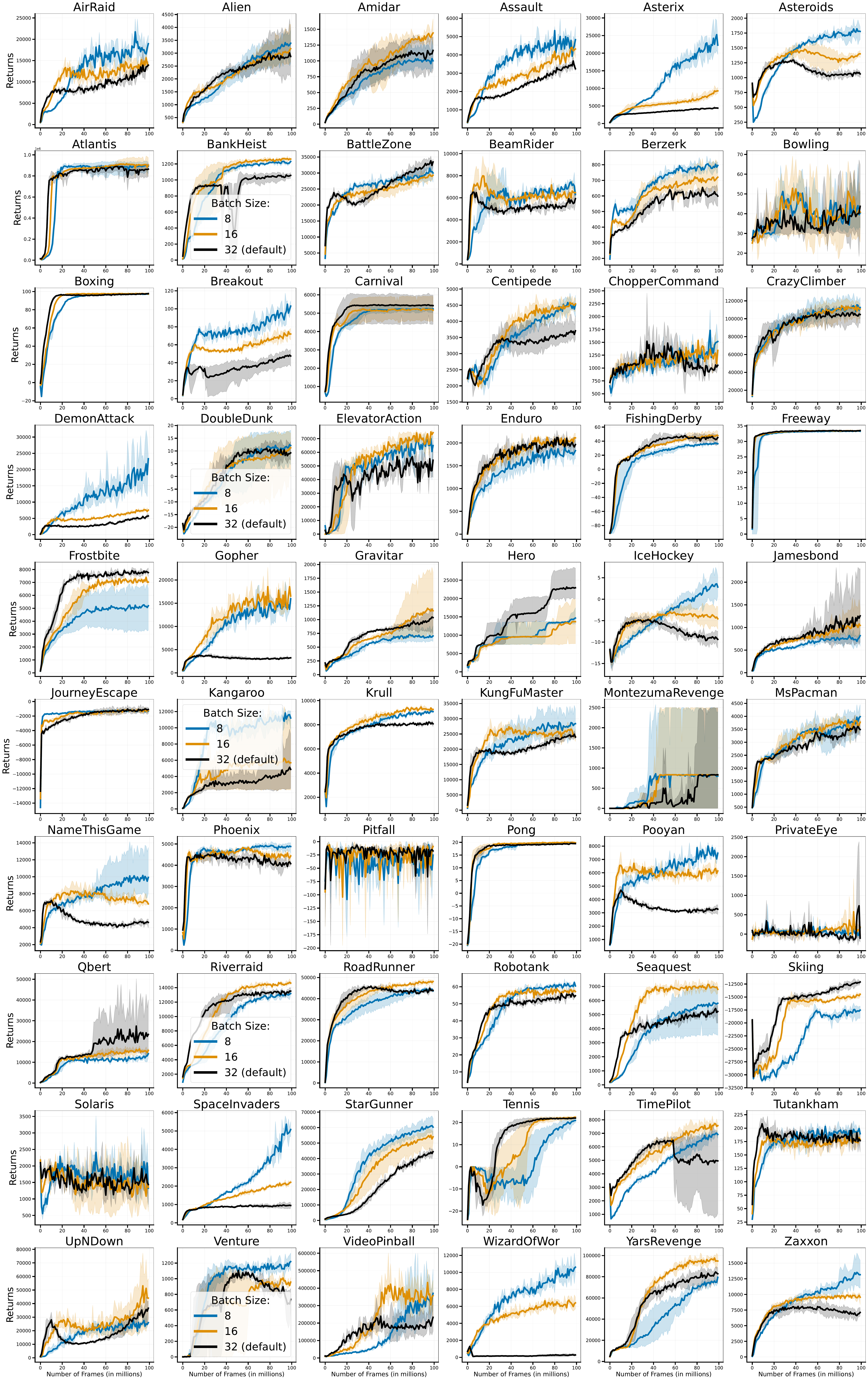}
    \caption{Training curves for QR-DQN agent. The results for all games are over 3 independent runs.}
    \label{fig:all_games}
\end{figure}

\section{Varying architectures}
\label{sec:appen_varying_architectures}

\begin{figure}[!h]
    \centering
    \includegraphics[width=0.835\textwidth]{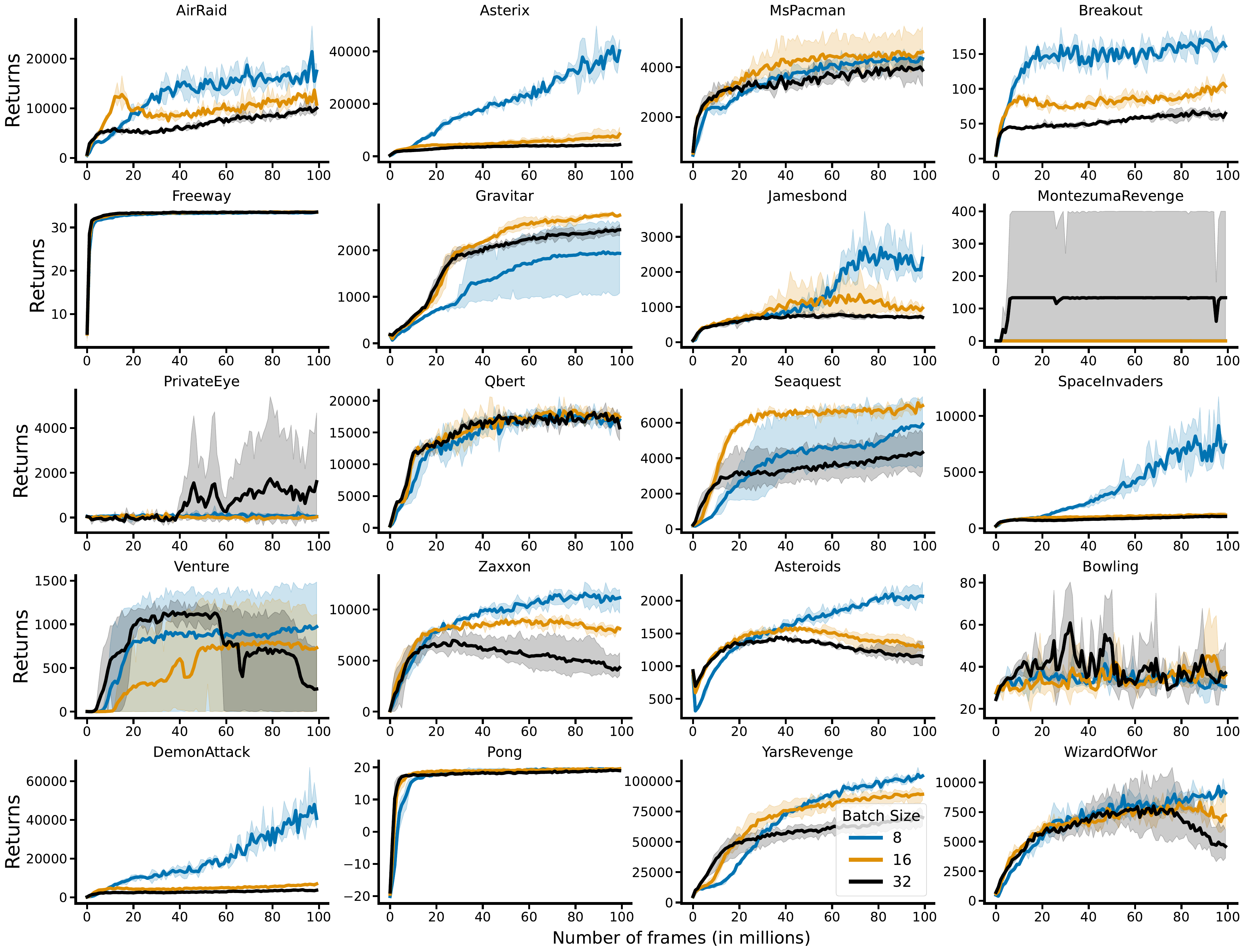}
    \caption{Evaluating the effect of CNNx4 to QR-DQN, learning curves for all games. Results averaged over 3 seeds, shaded areas represent 95\% confidence intervals.}
    \label{fig:cnnx4_allgames}
\end{figure}

\begin{figure}[!h]
    \centering
    \includegraphics[width=0.835\textwidth]{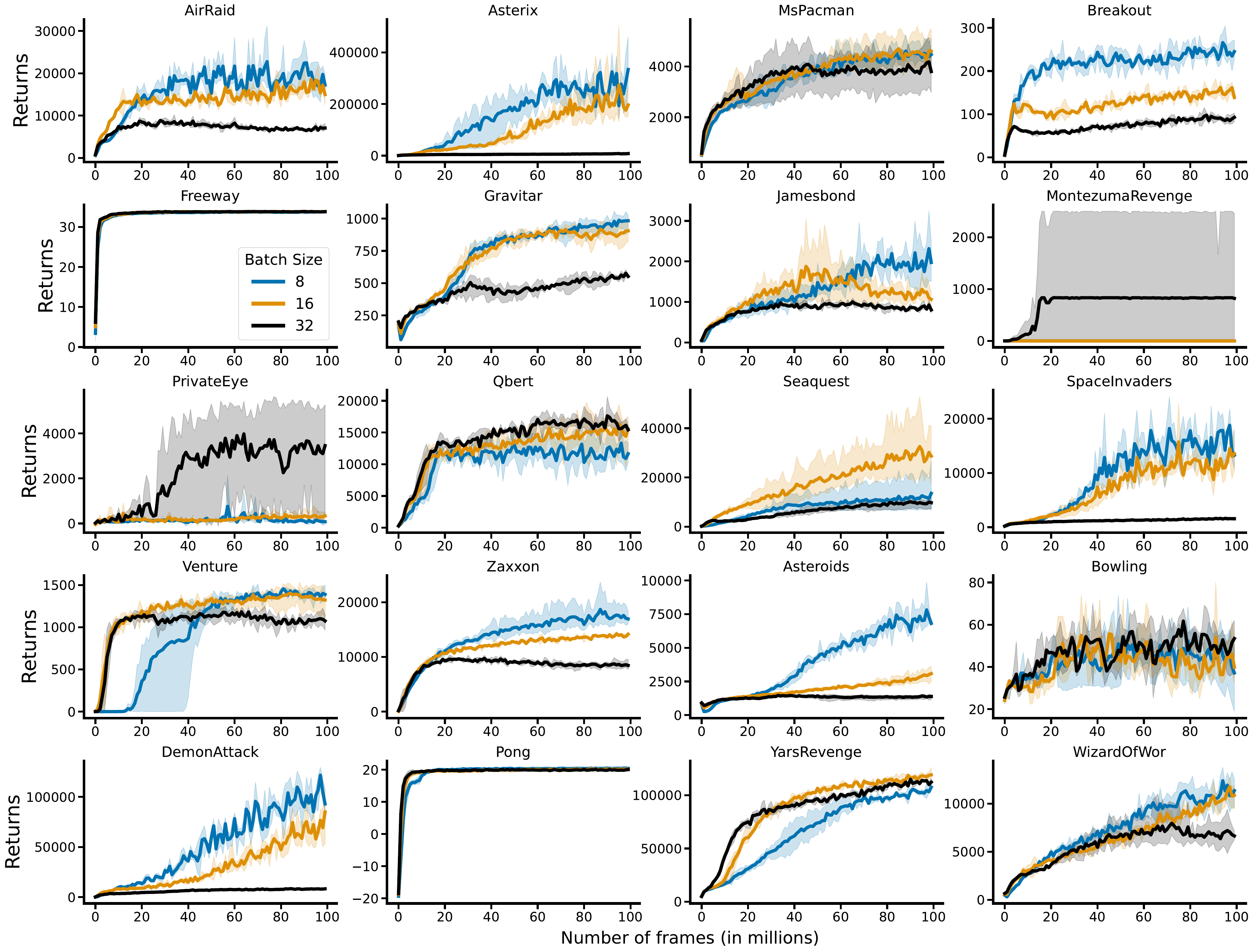}
    \caption{Evaluating the effect of Resnet to QR-DQN, learning curves for all games. Results averaged over 3 seeds, shaded areas represent 95\% confidence intervals.}
    \label{fig:resnet_allgames}
\end{figure}

\newpage
\section{Training Stability}
\label{sec:appen_training_stability}

\begin{figure}[!h]
    \centering
    \includegraphics[width=0.5\textwidth]{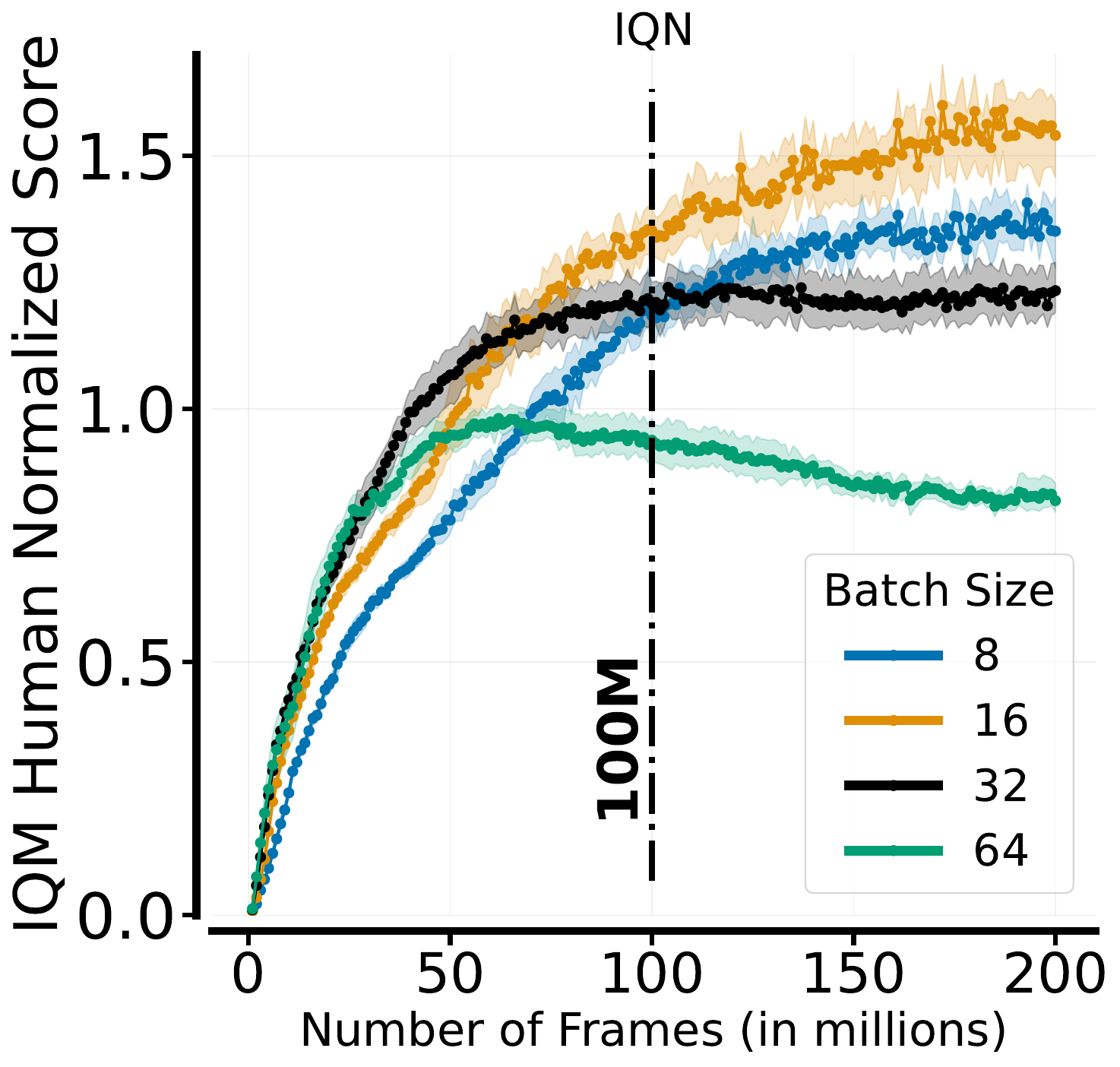}
    \caption{Measuring IQM for human-normalized scores when training for 200 million frames using IQN \citep{dabney18iqn}. Results aggregated over 20 games, where each experiment was run with 3 independent seeds and we report 95\% confidence intervals.}
    \label{fig:}
\end{figure}

\begin{figure}[!h]
    \centering
    \includegraphics[width=\textwidth]{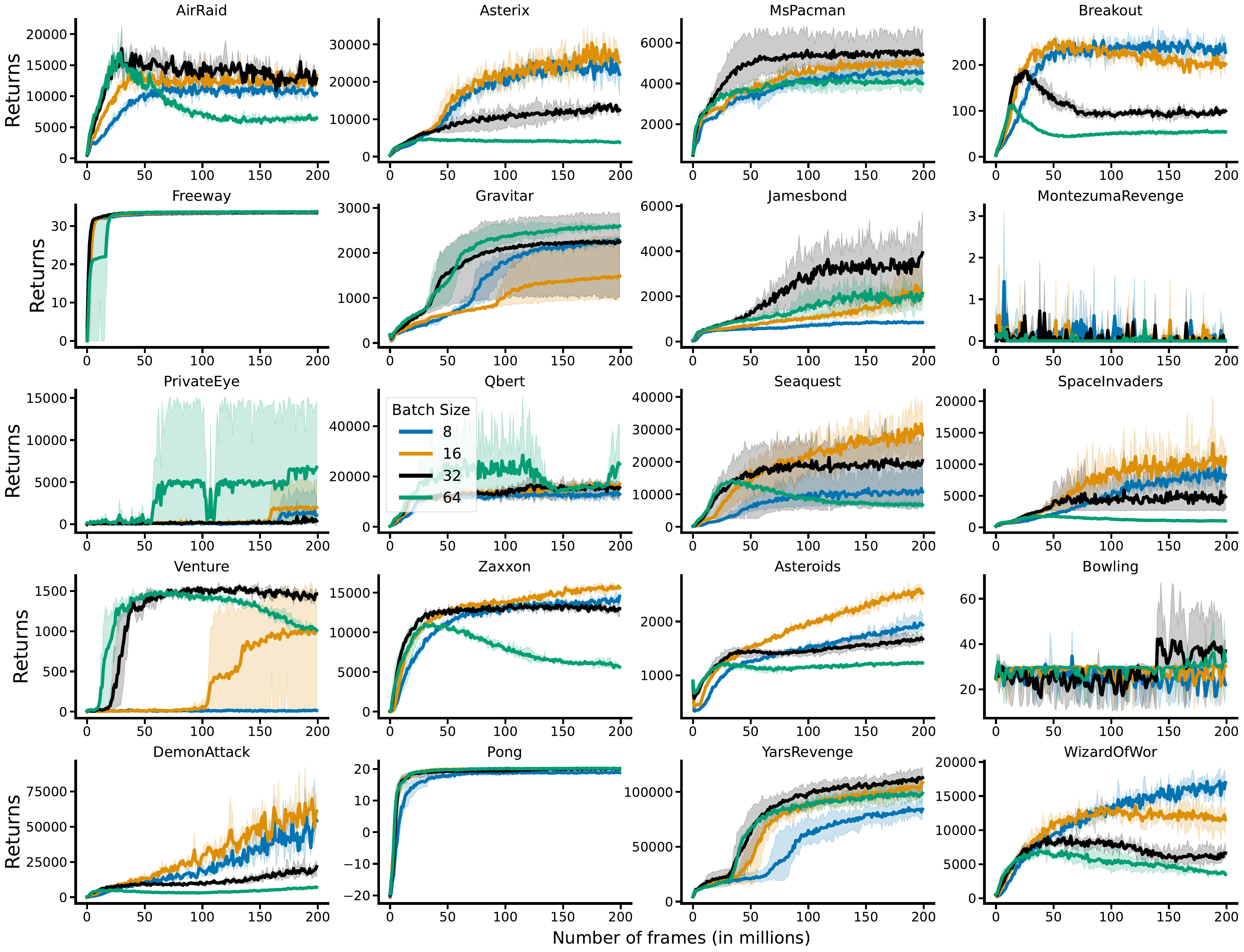}
    \caption{Learning curves for individual games, when trained for 200 million frames using IQN \citep{dabney18iqn}. Results aggregated over 3 seeds, reporting 95\% confidence intervals.}
    \label{fig:}
\end{figure}

\begin{figure}[!h]
    \centering
    \includegraphics[width=\textwidth]{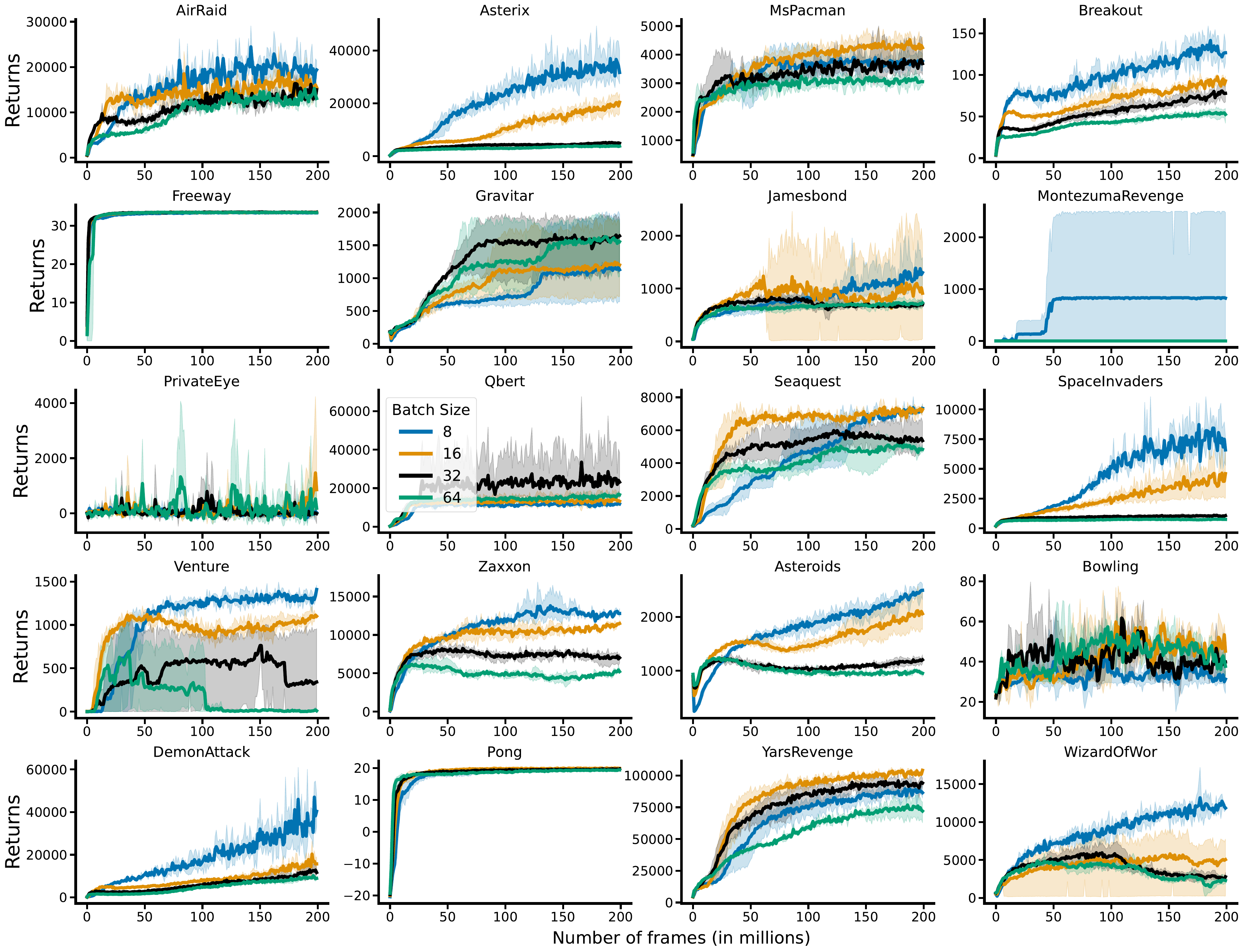}
    \caption{Learning curves for individual games, when trained for 200 million frames using QR-DQN \citep{dabney18distributional}. Results aggregated over 3 seeds, reporting 95\% confidence intervals.}
    \label{fig:}
\end{figure}
 

\end{document}